\documentclass[letterpaper, 10 pt, conference]{ieeeconf}      

\IEEEoverridecommandlockouts                              

\usepackage[nolist,nohyperlinks]{acronym}
\usepackage{amsmath} 
\usepackage{cases}
\usepackage{array}
\usepackage{amssymb}  
\usepackage[colorinlistoftodos]{todonotes}
\usepackage{hyperref}
\usepackage{cleveref}
\usepackage{multirow}
\usepackage{booktabs}
\usepackage[colorinlistoftodos]{todonotes}
\usepackage{textcomp}
\usepackage{gensymb}
\usepackage{siunitx}
\usepackage{url}
\usepackage{mathtools}
\usepackage{subcaption}
\usepackage{pgfplots} 
\usepackage{bm}  
\usepackage{soul}
\usepackage{dblfloatfix}
\usepackage{cite}
\usepackage{svg}
\usepackage{changes}

\newlength\fwidth
\newlength\fheight

\providecommand{\keywords}[1]{\textbf{\textit{Index terms---}} #1}

\title{\LARGE \bf
Design and Motion Planning for a Reconfigurable Robotic Base}

\author{Johannes Pankert, Giorgio Valsecchi, Davide Baret, Jon Zehnder, Lukasz L. Pietrasik, Marko Bjelonic, Marco Hutter  
\thanks{All authors are with the Robotic Systems Lab, ETH Zurich. Contact email: {\tt\small pankert@ethz.ch} ~ This work was supported by the Swiss Federal Railways (SBB), the Swiss National Science Foundation through the National Centre of Competence in Digital Fabrication (NCCR dfab), and by Innosuisse as part of project No. 51928.1.}
}

\begin{document}

\maketitle
\thispagestyle{empty}
\pagestyle{empty}

\begin{abstract}
A robotic platform for mobile manipulation needs to satisfy two contradicting requirements for many real-world applications: A compact base is required to navigate through cluttered indoor environments, while the support needs to be large enough to prevent tumbling or tip over, especially during fast manipulation operations with heavy payloads or forceful interaction with the environment.\\
This paper proposes a novel robot design that fulfills both requirements through a versatile footprint. It can reconfigure its footprint to a narrow configuration when navigating through tight spaces and to a wide stance when manipulating heavy objects. Furthermore, its triangular configuration allows for high-precision tasks on uneven ground by preventing support switches.\\
A model predictive control strategy is presented that unifies planning and control for simultaneous navigation, reconfiguration, and manipulation. It converts task-space goals into whole-body motion plans for the new robot.\\
The proposed design has been tested extensively with a hardware prototype. The footprint reconfiguration allows to almost completely remove manipulation induced vibrations. The control strategy proves effective in both lab experiment and during a real-world construction task.
\end{abstract}

\section{Introduction}
In the modern construction industry, there is enormous potential for automation. In contrast to the eight times increase in productivity of the manufacturing and agriculture industry, the productivity in the space of construction has largely stagnated \cite{TheMcKinsey}. There is a shortage of skilled manual labor, and the number of work-related accidents is high \cite{suva2020}.\\
Mobile Manipulators can be deployed on construction sites to work on tedious and dangerous tasks increasing the overall productivity \cite{hack2017mesh, helm2012mobile}. Still, existing construction robots are restricted to controlled environments and are mostly part of research labs.
We believe that mobility is a remaining challenge for today's construction robots that hinders widespread adoption in the industry.
Specifically, a mobile robot has to navigate through tight spaces while supporting heavy loads like a large manipulator equipped with power tools.
Some construction tasks, such as drilling or grinding, require high positional accuracy of the robot's end-effector. Uncontrolled wobbling motion due to switching supports of the robotic base reduces the achievable accuracy.
None of the existing solutions for construction robots perform well in all said categories.
In this work, we present a novel, reconfigurable mobile robot that can navigate tight spaces, support heavy loads and enable highly accurate manipulation.
Its key feature is the ability to reshape its support polygon autonomously.
The proposed design allows for omnidirectional mobility thanks to individually steerable wheels.
In contrast to platforms with mecanum drives, the robot cannot travel in all directions instantaneously, making the control challenging.
We have developed a receding horizon control strategy that allows for smooth trajectory tracking by incorporating the wheel constraints. The controller can also reconfigure the support polygon while driving allowing fast task execution.
\begin{figure}[t]
\centering
\includegraphics[width=\columnwidth]{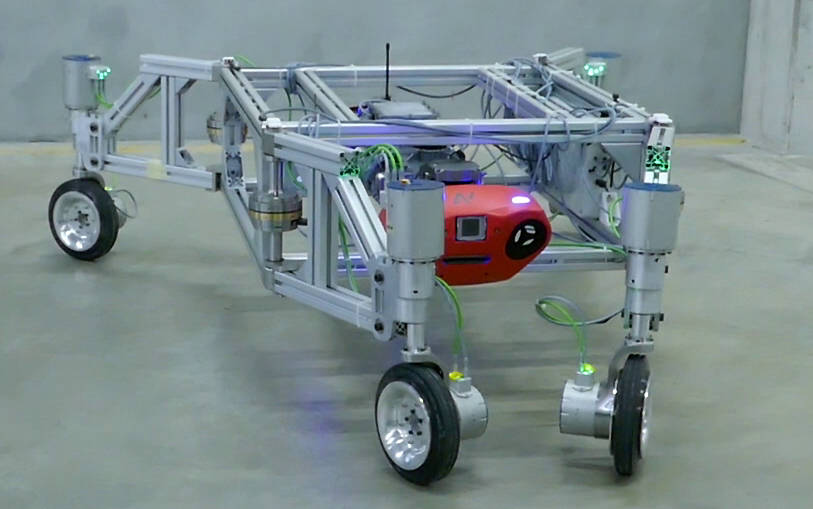}
\caption{Hardware prototype of the reconfigurable mobile base. The robot is in \emph{A-configuration} which is the most suitable for precise manipulation tasks.}
\label{fig:teaser}
\end{figure}

\subsection{Related Work}
In the following section, we review state of the art in mobile robot design and motion planning for reconfigurable mobile manipulators:
\subsubsection{Hardware Designs}
In road-going and offroad vehicles such as the \emph{AMBOT GRP 4400} \cite{WheeledCompany}, additional suspension elements at the wheels solve the problem of the tipping motion due to the switching of support polygons. Those suspensions are typically spring-damper systems or air-filled tires, which cause the wheels to maintain contact.
However, the suspension system still allows for a movement of the frame or body of the vehicle itself, which hinders precise manipulation tasks. 

Suspension systems can also have an active component for stabilizing the vehicle's body, like the linear actuators used in \emph{WorkPartner} \cite{Halme2000HybridMachine}, or the hydraulic suspension systems employed by Lotus \cite{Milliken1988ActiveSuspension} and Williams \cite{2017PatrickGreat} in Formula 1. These systems can mitigate base tipping but require constant actuation, resulting in additional actuators, cost, and power draw.

The problem of switching support polygons can also be overcome by providing only one triangular support. This approach results in three-wheeled solutions deployed on pioneer robots  \cite{Pioneer3-DX} or the \emph{Robotnik RB-1} mobile bases \cite{RB-1Robotnik}.
However, those robots cannot extend their maximum footprint beyond what the platforms are constrained to.

Another approach to increase the total support polygon area on large construction machinery such as cranes is using outriggers like the ones from \emph{autocrane} \cite{OutriggersCrane}, described for example in a patent \cite{JanBirgerPalmcrantz1973SupportCranes} or foldable ones employed by \emph{Husqvarna}'s demolition robots \cite{RemoteProducts}. The extendable supports of outriggers can only be used while the vehicle is stationary, restricting its mobility. 

Wheeled-walking excavators change their support polygon dynamically for safety reasons \cite{Mandrovskiy2018OptimizingStability}. Such machines require additional actuators and are more costly in draw and weight.

The support polygon of a mobile robot base can also be changed through the use of articulated legs \cite{Kamedula2020ReactiveRobot}. With force-controllable legs, the ground reaction forces can even be balanced actively \cite{Hutter2016ANYmalRobot}. Again, such solutions require additional actuation.

Another solution to the tip-over-stability problem is adding a counter mass to the manipulator base. This measure decreases the effect of the manipulation action to the center of mass position of the robot. Boston Dynamics’ Stretch \cite{StretchDynamics}, for example, weighs about 1250 kg \cite{BostonPalletizing}. A large mass decreases the agility of the robot base and the number of places where it can operate while increasing effort for transportation and power draw.

\emph{ROAMeR}, a robotic research platform utilizes its actuated reconfigurable legs to provide access to constrained spaces \cite{Fu2014TheROAMeR}.
Another example for reconfigurable legs with wheels mounted on them is IRIM Lab's Agile Omnidirectional Mobile Robot with Gravity Compensated Wheel-Leg Mechanisms for Human Environment \cite{Yun2021DevelopmentEnvironments}. Compared to ROAMeR, the IRIM Lab's robot not only reconfigures its arms to be at a different angle in the ground plane but also uses them to lift its main body off the ground. 
\subsubsection{Whole-Body Motion Planning and Control}
\emph{ROAMeR} uses a purely reactive hierarchical controller where the navigation task is the primary objective, and the reconfiguration task is projected into the nullspace.
Nullspace projections are used as well in an optimization-based whole-body controller for wheeled legged robots \cite{anymal_on_wheels}.
Other works completely separate control for navigation and reconfiguration \cite{Yun2021DevelopmentEnvironments}.
The approaches mentioned above do not plan motions for all degrees of freedom but separate the generation of reference trajectories and whole-body control.
Because of the non-holonomic constraints of our robot's wheels and the unactuated leg joints, planning and control require a unified solution.

Sampling-based planners such as RRTs have been shown to be suitable for whole-body planning for mobile manipulators \cite{burget_whole-body_2013}. While they can find collision-free trajectories that respect the given constraints, sampling in high-dimensional configuration space is slow.
Planning times in the order of seconds do not allow for continuous re-planning, which is necessary for disturbance rejection during dynamic maneuvers.

\emph{CENTAURO} robot uses A* to perform a graph search for combined navigation and reconfiguration \cite{raghavan_variable_2019}. The method works well in the presented application for planar navigation with obstacle negotiation through reconfiguration.
It is doubtful that it scales to higher dimensional problems, though.
For manipulation tasks, the degrees of freedom of a robotic arm have to be considered.

In recent years, reinforcement learning has become a popular approach for controlling mobile manipulators. Some works separate the locomotion and the manipulation problem,  only learning the locomotion policy \cite{honerkamp2021learning, alma_learning}.
Other works learn both navigation and manipulation but rely on existing low-level controllers and use a symbolic action space \cite{sun2022fully}.

Trajectory optimization methods like Differential Dynamic Programming (DDP) scale well to high dimensional problems allowing for fast replanning, and have successfully been deployed on mobile manipulators \cite{alma_mpc, giftthaler_efficient_2017, pankert2020perceptive}.
Building upon our previous work \cite{pankert2020perceptive}, we show how DDP can be used in a receding horizon fashion as a Model Predictive Control (MPC) strategy to plan motions for a reconfigurable robot reactively.
\subsection{Contributions}
This work presents the design and control of a novel mobile manipulator for construction applications.
Through reconfiguration, it can navigate tight spaces and enable precise manipulation.
It can change the footprint without additional actuated degrees of freedom.
The developed motion planning module is released as open-source software. It can be used on both reconfigurable and fixed-base swerve steering robots and provides significant advantages over other freely available swerve steering controllers.
\section{Design}
\label{sec:methods}
We postulate that a mobile robot should have a stable stance to perform precise manipulation tasks.
This section motivates the design choices by first discussing how a stable stance can be defined. We then show how our concept of reconfigurable legs can ensure stability and finally demonstrate how the idea translates to a hardware design. 
\subsection{Stability}
The convex hull of all contacts of a mobile robot is called the support polygon (SP).
The Zero Moment Point (ZMP) is the point at which the sum of all tipping moments induced by gravity, inertia, and external wrenches is zero.
A mobile robot does not fall over when its ZMP is inside the SP \cite{zmp}.
This definition of stability has been used in the legged robotics community for many years to design dynamic-walking controllers \cite{shigemi2018asimo}.\\
However, for precise mobile manipulation, we find that this stability criterion is necessary but not sufficient.
We observe that rigid mobile bases with four desired contact points (i.e. wheels) typically have their weight distributed only to three contacts.
This happens because the four contacts do not lie in the same plane, either due to imperfections in the mechanical design or uneven ground.
In case of a perfectly rigid mobile base and rough terrain, one contact is in the air. In case of compliance at the contact points or in the system itself, the contact force would be significantly lower for one contact than for the other three.
These three active contact points form the robot's active support triangle (ST).
The ZMP moves as the manipulator operates, and the external wrench at the end-effector changes during manipulation actions.
When the ZMP leaves one ST, stays in the overall SP but enters another ST, the robot will not fall over, but the shift in support can cause wobbling.
A small change in base orientation amplified with the length of the manipulator may produce a significant displacement of the end-effector.
During high-accuracy manipulation tasks, such an unexpected shift could, for example, damage a workpiece.
We, therefore, conclude that for mobile manipulation, two criteria are essential for stability:
\begin{itemize}
    \item The ZMP in SP criterion guarantees that the robot does not fall over.
    \item The ZMP in ST criterion prevents the base from wobbling.
\end{itemize}
\subsection{Leg Configurations}
\label{ss:configurations}
Our proposed robot design has four legs that are attached to the main body with passive joints as shown in \cref{fig:teaser}.
The legs are equipped with steerable wheels that can be used both for locomotion and to change the joint positions of the passive leg joints.
When the leg joint positions do not have to change during operation, the passive joints can be locked with magnetic brakes. This increases the stiffness of the robot.

\begin{figure}[tb]
	\centering
	\begin{subfigure}[t]{0.32\linewidth}
		\includegraphics[width=1.0\linewidth]{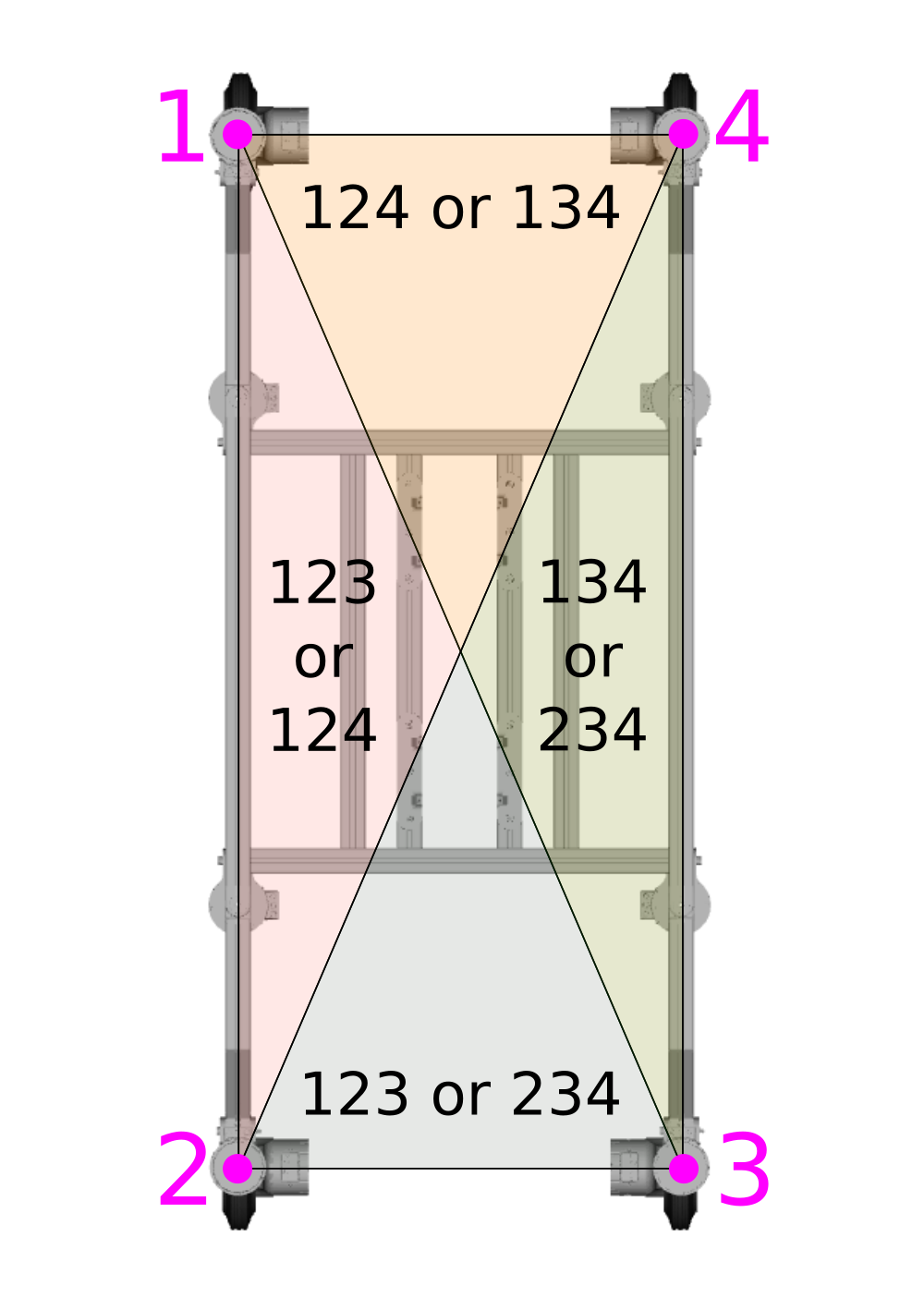}
		\caption{H-Configuration}
		\label{fig:H_conf}
	\end{subfigure}
	\hfill
	\begin{subfigure}[t]{0.32\linewidth}
		\includegraphics[width=1.0\linewidth]{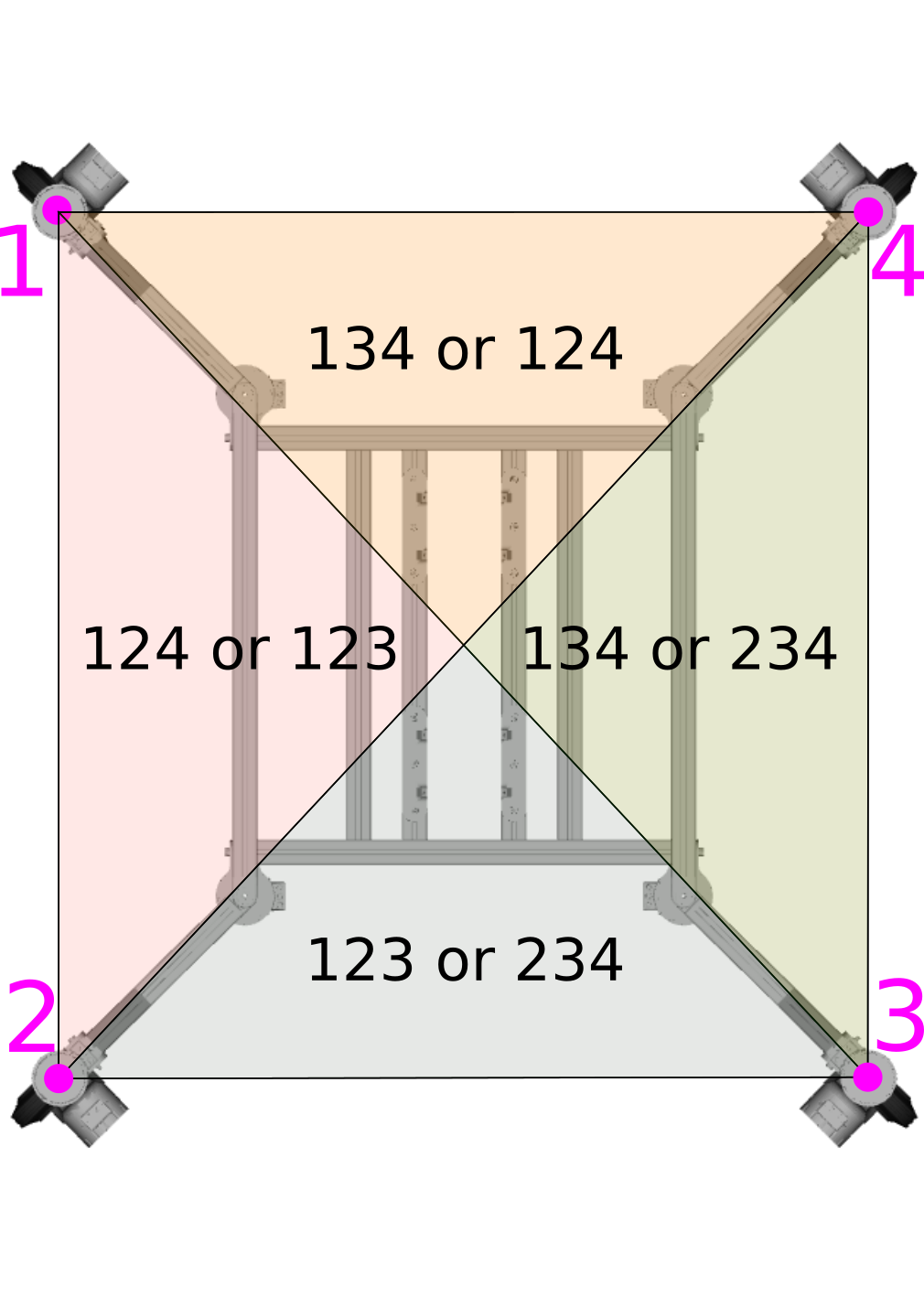}
		\caption{X-Configuration}
		\label{fig:X_conf}
	\end{subfigure}
	\hfill
	\begin{subfigure}[t]{0.32\linewidth}
		\includegraphics[width=1.0\linewidth]{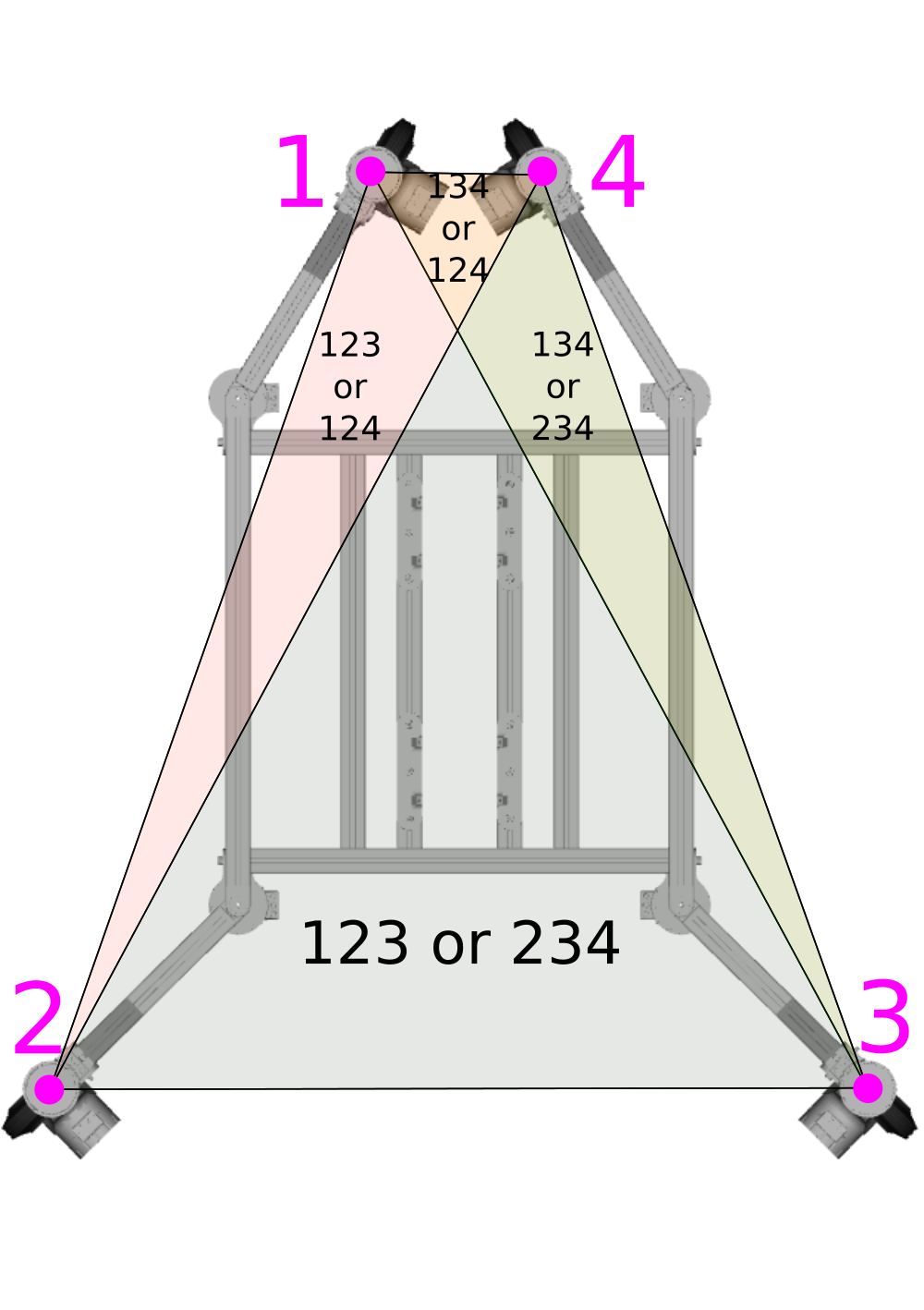}
		\caption{A-Configuration}
		\label{fig:A_conf}
	\end{subfigure}
	\caption{Visualization of the contact points $\{1,2,3,4\}$ and possible support triangles. The support triangles are named by a tripled of vertices (i.e. 124). The colored sections show where two different support triangles overlap. The base might wobble when the ZMP transitions from one triangle to another.
	}
	\label{fig:configurations}
\end{figure}

We call a set of joint positions for the legs \emph{configuration}.
Even though the mechanism can steer the joints into arbitrary positions, we analyze three distinct configurations in this work: \emph{H-configuration}, \emph{X-configuration}, and \emph{A-configuration}.
\Cref{fig:configurations} shows different configurations as well as the possible STs. Two connected colored areas are one ST. Each colored area is the union of two different STs. We report the areas of the SPs and the largest unions of STs for all configurations in \cref{tab:areas}.

The X-configuration has a large SP and is, therefore, the safest option to handle heavy payloads or large external forces according to the ZMP in SP criterion.\\
We propose to operate the robot in A-configuration when high precision is needed. In this asymmetric configuration, two support points are close to each other, and two other points are further apart.
This creates two large STs that are overlapping.
During nominal operation, the ZMP stays in the union of both STs.
This prevents switching from one ST to another and therefore prevents wobbling.\\
The \emph{H-configuration} has the narrowest footprint and is useful for navigation in confined spaces such as passing through doors.
\begin{table}[tb]
\caption{Sizes of the support polygons (SP) and the largest unions of support triangles (ST) for the three configurations.}
\label{tab:areas}
\begin{tabular}{c|cc}
\textbf{Configuration} & \textbf{SP Area }$[\si{\square\meter}]$ &  \textbf{Largest Union of ST Area $[\si{\square\meter}]$} \\
H & 1.16 & 0.29 \\
X & \bf{1.81} & 0.45 \\
A & 1.17 & \bf{0.76}
\end{tabular}
\end{table}
\subsection{Hardware Design}
\label{ss:hw}
We built a prototype to validate the proposed mobility concept. To define design specifications, we decided to focus on applications in partially finished buildings that heavy machinery cannot access but where tasks are still labor intense. Examples of these applications are plastering, grinding, drilling and chiseling. Three major requirements emerged: a payload of \SI{150}{kg}, ingress protection from dust and water, and the possibility of fitting through doors. The robot's main body is composed of a structural frame of aluminum profiles connected to the legs. The structure also houses a sealed unit containing computers, battery, and communication modules. The reconfigurable legs, shown in \cref{fig:frames}, are also made out of aluminum profiles.
To properly lock the position of the leg, a shaft with a brake, Mönninghoff 560.21.4.5 \cite{kupplungshersteller}, is connected to the rotation axis. When the brake is active, the rotation is locked, and up to \SI{200}{Nm} can be sustained. The positions of the legs are read by four on-axis encoders \cite{seeedstudio}. The steerable wheels are powered by two ANYdrives, one for the driving torque and one for steering. To perform manipulation tasks, we mounted either a MABI Speedy 12 or a Universal Robots UR10 arm on the frame. 

\section{Motion Planning and Control}
In this section, we will describe the control framework of the reconfigurable base. We also describe how the base twist can be estimated from wheel odometry.
\subsection{Model Predictive Control}
\label{ss:mpc}
\begin{figure}
	\centering
	\includegraphics[width=1.0\linewidth]{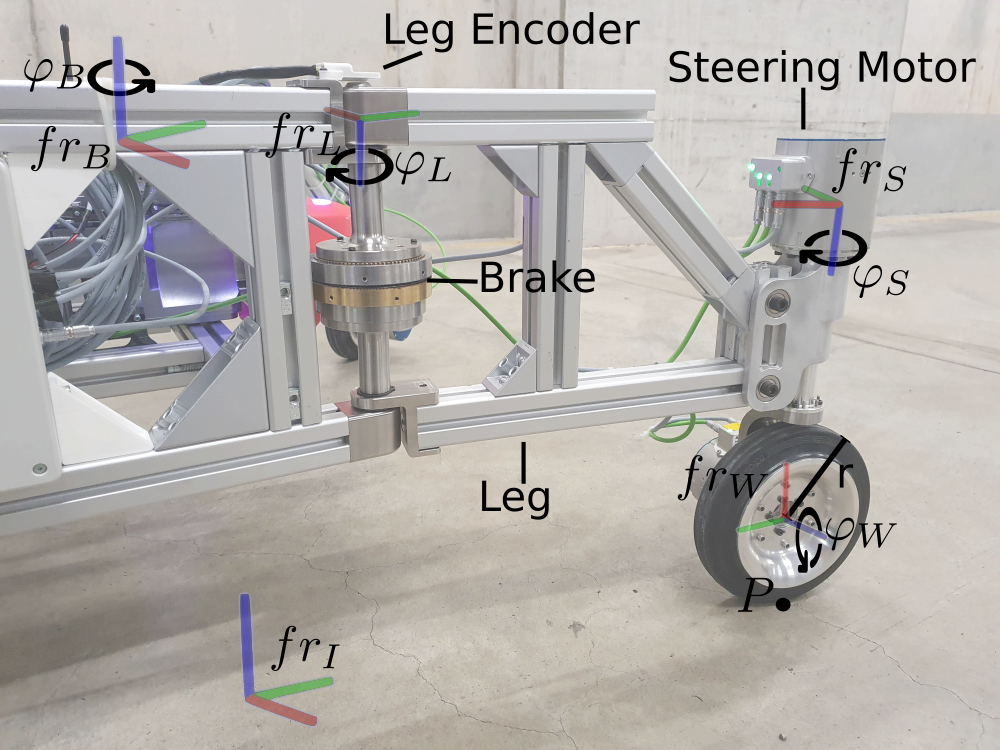}
	\caption{Picture of a single leg. The leg is attached to the main body with a passive joint. The joint can be locked with a brake. An encoder measures the joint position.\\
	Frame conventions: $fr_{\{I,B,L,S,W\}}$ are the inertia, body, leg, steer and wheel frames. The point $P$ is the contact point between wheel and ground. $\varphi_{\{B,L,S,W\}}$ are the respective generalized coordinates.}
	\label{fig:frames}
\end{figure}

A model predictive control (MPC) setup is adopted to plan and execute motion for the mobile robot.
We distinguish a \emph{tracking mode}, where the base follows a given path, and a \emph{whole-body mode}, where the end-effector tracks the desired trajectory.
The ILQR-MPC implementation of the OCS2 framework \cite{farshidian2017efficient} is used to solve the MPC problem. The end-effector tracking objective for whole-body motion planning and the base motion model formulation follow the ideas of our previous work \cite{pankert2020perceptive}.
The MPC uses a kinematic model. The system state $\bm{x}$ consists of the floating base pose $\bm{x_b} = [\bm{t_b}, \bm{q_b}]^T$, the leg angles $\bm{\varphi_L}$, the steering angles $\bm{\varphi_S}$, the wheel angles $\bm{\varphi_W}$ and the arm joint positions $\bm{x_{arm}}$.
$\bm{t_b}$ is the base position in inertia frame. $\bm{q_b}$ is the base orientation in inertia frame parameterized with a quaternion.
The states of the brakes $\bm{\lambda}$ take discrete values. Zero for a closed brake and one when a brake is open. 
\Cref{fig:frames} shows the frame conventions for one leg.
The rigid body transformations can be extracted from the CAD design of the assembly.
The system dynamics are governed by a flow map and a set of constraints:
\begin{align}
	\bm{\dot{x}} &=  \begin{pmatrix}
	\bm{\dot{x_b}},
	\bm{\lambda}\odot\bm{\omega_L},
	\bm{\omega_S},
	\bm{\omega_W},
	\bm{\omega_{arm}},
	\end{pmatrix}^T\\
    \bm{\dot{x_b}} &=
    \begin{pmatrix}
    \bm{q_b}\cdot[v_x, v_y, 0]^T, \bm{q_b} \boxdot 0.5k\omega_b
    \end{pmatrix}^T\\
	\bm{v_p}^{fr_I}&=\bm{0}\label{eq:wheel_constraint}
\end{align}
$\odot$ is the element-wise product, $\cdot$ the quaternion vector product, $\boxdot$ the Hamilton quaternion product, and $k$ the third quaternion imaginary unit.
$(v_x, v_y, \omega_b)$ is the planar base twist, and $\bm{\omega_{\{L, S, W\}}}$ are the angular velocities of the respective joints.\\
The constraints \cref{eq:wheel_constraint} ensure that the planned instantaneous velocity of the wheel contact points $P$ is zero in the inertial frame.
They prevent the wheels from sliding sideways and slipping.
The constraints are implemented by transforming the base twist to the steering frames and enforcing the following identities with $r$ being the wheel radius:
\begin{align}
	\bm{v_x}^{fr_S} &= r\bm{\omega_W}\\
	\bm{v_y}^{fr_S} &= \bm{0}
\end{align}
The joint position and velocity limit inequality constraints are added as soft constraints and implemented with Relaxed Barrier Functions (RBF) as part of the cost function since our DDP solver cannot handle hard inequality constraints.

Base or end-effector trajectories are followed with quadratic cost functions that penalize the deviation of the current pose to the desired pose over the time horizon.
The desired leg positions are tracked with a quadratic cost term as well.
All cost terms are summed up and integrated over the time horizon:
\begin{align}
J(x, u, t)&=\int_{t}^{t+T} L(\mathbf{x}, \mathbf{u}, \tau)d\tau\\
L(\mathbf{x}, \mathbf{u}, \tau) &= L_{tr} + L_{conf} + L_{sc} + \bm{u^T}\bm{R}\bm{u}\\
L_{tr}(\bm{x}, \bm{\tau}) &= ||\bm{x_b}(\tau) - \bm{\hat{x}_b}(\tau)||^2_2\\
&~~~~\text{or}~ ||\bm{x_{ee}}(\tau) - \bm{\hat{x}_{ee}}(\tau)||^2_2\nonumber\\
L_{conf}(\bm{x}, \bm{\tau})&=||\bm{\varphi_L}(\tau) - \bm{\hat{\varphi}_L}(\tau)||^2_2\\
L_{sc} &= \text{RBF soft contraints}
\end{align}
The vector $\bm{u}=(\bm{v}, \omega_{\varphi_B}, \bm{\omega_{\varphi_L}}, \bm{\omega_{\varphi_S}}, \bm{\omega_{\varphi_W}}, \bm{\omega_{\varphi_{arm}}})^T$ is the control input in the DDP problem.
Only the $\bm{\tilde{u}}=(\bm{\omega_{\varphi_S}}, \bm{\omega_{\varphi_W}}, \bm{\omega_{\varphi_{arm}}})^T$ terms are used as actuation commands for the robot hardware.
The control inputs to the floating base and the legs are auxiliary inputs to simplify the formulation of the system dynamics and constraints.\\
The time horizon of the MPC is $T=\SI{10}{\second}$ and the update rate for the optimization is $\SI{50}{\hertz}$.
In this work, reconfiguration events are triggered externally, and reference trajectories for the leg positions are provided. Alternatively, the references could also be omitted. The leg positions would then be determined by the MPC trying to optimize for additional objectives such as environment collision avoidance or ZMP in SP/ST stability.
\subsection{Odometry}
\label{ss:odom}
Wheel odometry estimates the robot's base motion given the recorded wheel velocities.
This estimation is typically not very accurate for skid steer or tracked vehicles since the wheels or tracks slip during turning maneuvers, but for a swerve steering platform, the state estimation can rely on the wheel velocity information.
The base twist can be estimated by first computing how the base and leg motions affect the velocities of the wheels. We can then use least squares to solve for the base twist.\\
The velocity of a steering frame is a superposition of linear base velocity and the angular velocities of the base and leg:
\begin{align}
\bm{v}^{fr_B}_S =&~ \bm{v}^{fr_B}_B + \bm{t}^{fr_B}_S \times \bm{\omega}^{fr_B}_B \label{eq:odom}\\
 &~+(\bm{t}^{fr_B}_S - \bm{t}^{fr_B}_L) \times \bm{\omega}^{fr_B}_L \nonumber\\
\bm{v}^{fr_B}_S =&~\bm{R}^{fr_B}_{fr_S} \cdot \bm{v}^{fr_S}_S = \bm{R}^{fr_B}_{fr_S} \begin{pmatrix}
\omega_w \cdot r \\
0
\end{pmatrix} \label{eq:steer_vel}
\end{align}
All velocities are relative to the inertia frame rotated by the respective frame from the superscript. The translation vectors $\bm{t}_{S,L}$ are expressed in base frame. \Cref{fig:frames} shows the frame conventions.
We can substitute $\bm{v}^{fr_B}_S$ with \cref{eq:steer_vel}, stack the equations of all the legs and solve for the base twist $(\bm{v}^{fr_B}_B, \bm{\omega}^{fr_B}_B)$ with Least Squares since \cref{eq:odom} is linear in the base twist.
\subsection{Open Source ROS-Control Implementation}
\label{sec:open_source}
The MPC implementation and the odometry module are released as open-source software\footnote{\url{https://github.com/leggedrobotics/swerve_steering}}. The ROS Control framework is used to interface with robotic hardware or a simulator \cite{ros_control}. We release the code for both reconfigurable robots and standard swerve steering mobile bases with fixed footprints. The controller brings significant improvements compared to the default \texttt{four\_wheel\_steering\_controller} since this controller can only track linear motion in x-direction and rotations but no sideways motion \cite{FourWheelSteering}.
\section{EXPERIMENTS}
\label{sec:experiments}
In this section, we validate the robot's performance. Firstly we show the impact of the robot's configuration on its support polygon and stability. Then the MPC capabilities are evaluated for base tracking and whole-body control mode. Finally, we show how the robot's features allow for successful deployment and task execution in a building construction environment.  
\subsection{Base Stability}
\label{ss:stability}
\begin{figure}[t]
	\centering
	\begin{subfigure}[t]{0.49\linewidth}
		\includegraphics[width=1.0\linewidth]{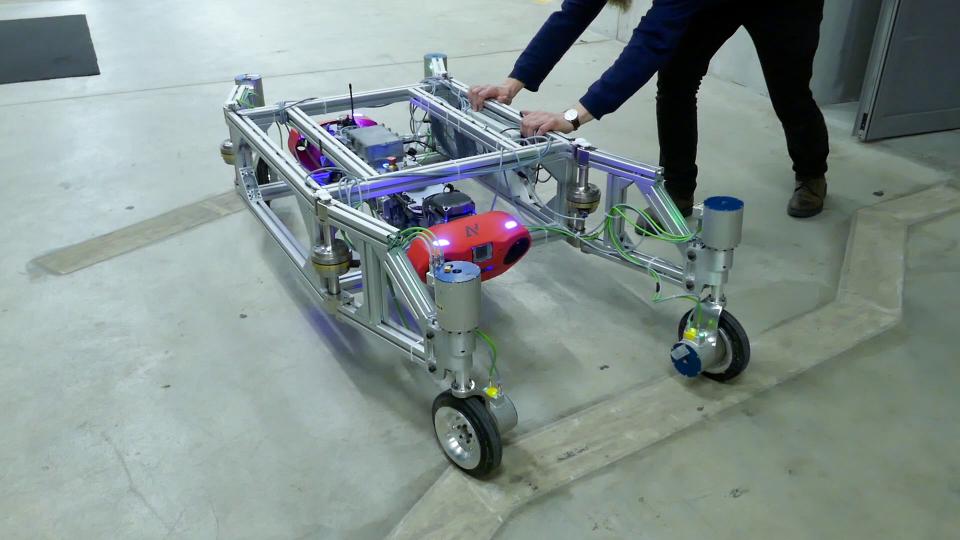}
		\caption{Brakes off}
		\label{fig:brakesOFF}
	\end{subfigure}
	\hfill
	\begin{subfigure}[t]{0.49\linewidth}
		\includegraphics[width=1.0\linewidth]{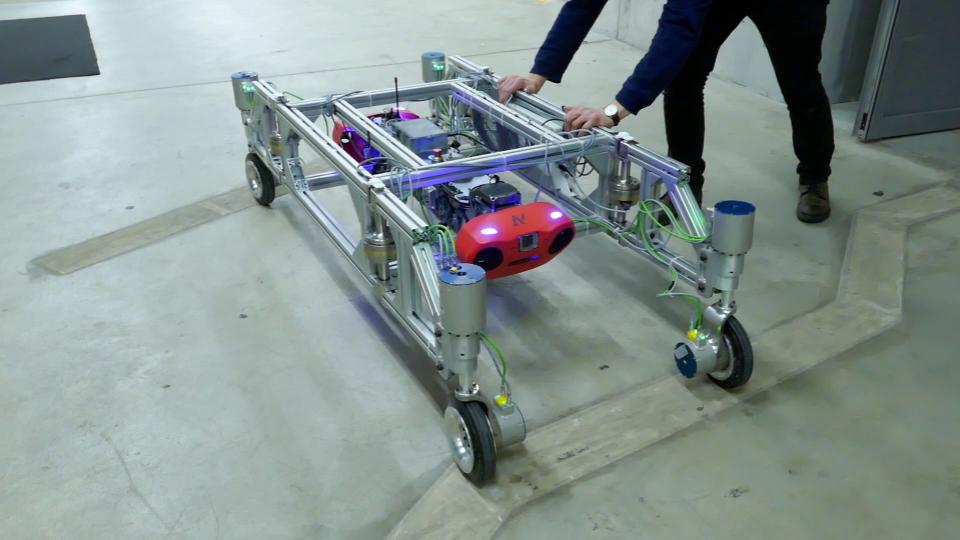}
		\caption{Brakes on}
		\label{fig:brakesON}
	\end{subfigure}
	\caption{Effect of brakes in H configuration. The reconfigurable base can counteract the disturbance when the brakes are engaged, and the legs do not move. The base cannot resist side pushes when the brakes are disengaged.}
	\label{fig:brakes}
\end{figure}

\begin{figure}[tb]
	\centering
	\begin{subfigure}[t]{0.2\linewidth}
		\includegraphics[width=1.0\linewidth]{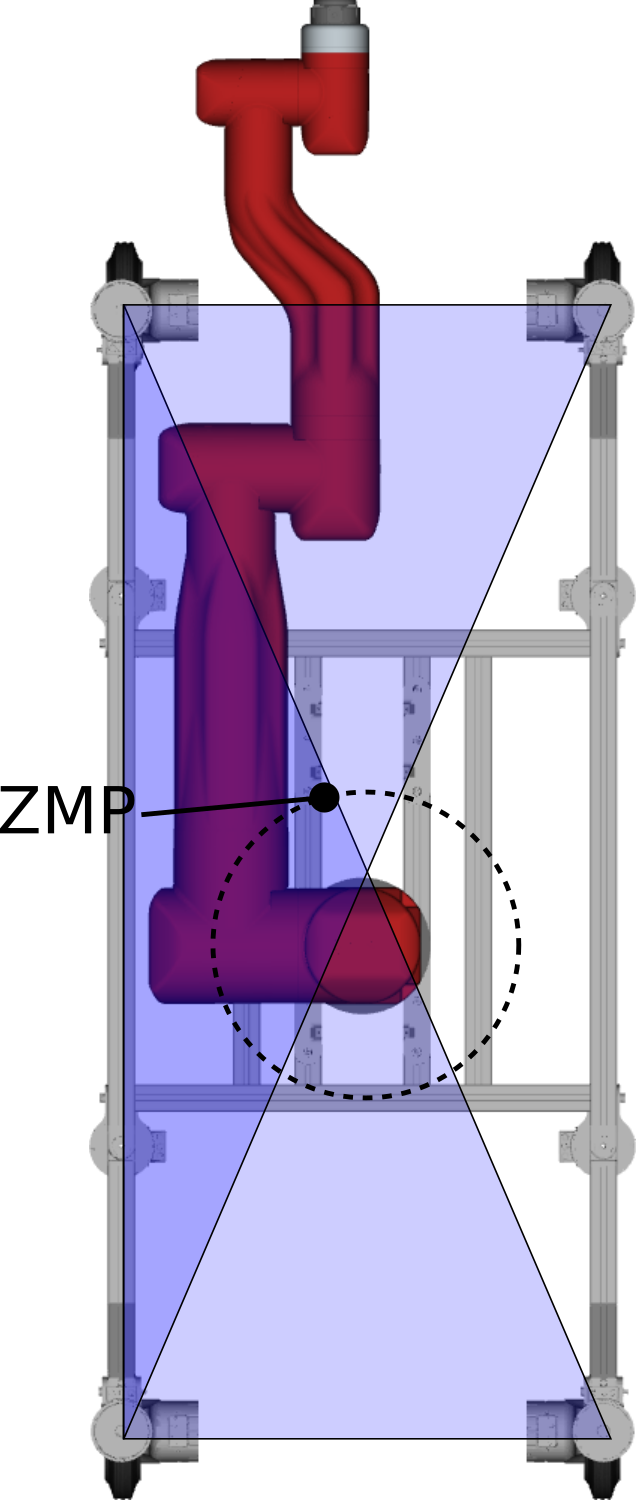}
		\caption{\emph{H-conf.}}
		\label{fig:H_conf_arm}
	\end{subfigure}
	\hfill
	\begin{subfigure}[t]{0.32\linewidth}
		\includegraphics[width=1.0\linewidth]{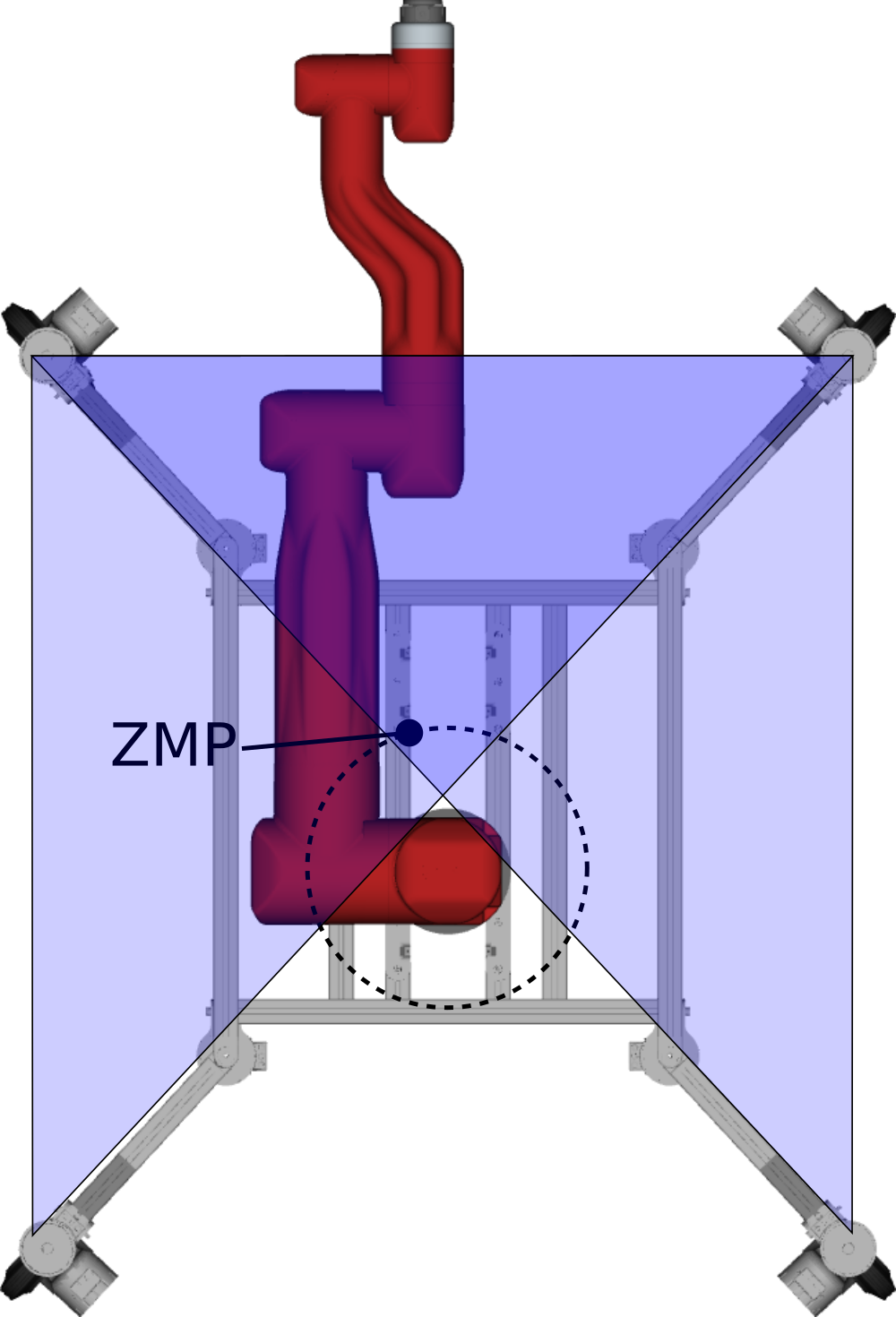}
		\caption{\emph{X-conf.}}
		\label{fig:X_conf_arm}
	\end{subfigure}
	\hfill
	\begin{subfigure}[t]{0.32\linewidth}
		\includegraphics[width=1.0\linewidth]{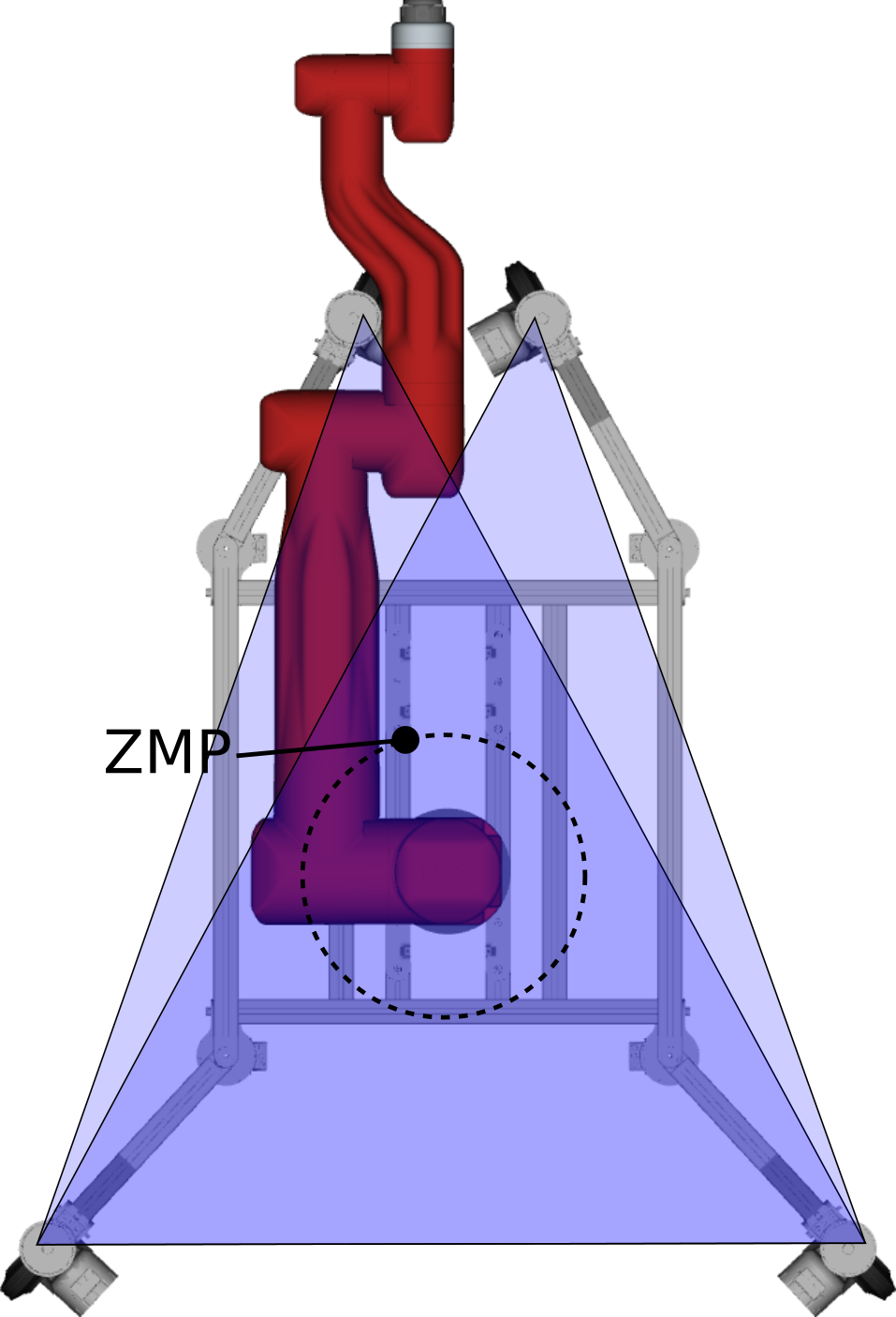}
		\caption{\emph{A-conf.}}
		\label{fig:A_conf_arm}
	\end{subfigure}
	\caption{Top view of the base with the robotic arm mounted in three different configurations. The black dashed circle shows qualitatively how the ZMP move when the arm turns. The dark blue areas are part of the active ST. The light blue areas may either be active or not active. When the ZMP transitions into an inactive support triangle, the base wobbles as shown in \cref{fig:angularVelocities}. If the robot is in \emph{A-configuration}, as demonstrated in \cref{fig:A_conf_arm}, the ZMP remains in the same ST, and the base will not wobble.
	}
	\label{fig:overlayCOM}
\end{figure}

We evaluated the robot stability in different configurations and against various disturbances.

The brakes are needed to fix the position of the legs. We tested their efficacy with the robot in H configuration when it is in a singular position and cannot resist forces perpendicular to the legs. \Cref{fig:brakes} shows the reduced displacement caused by a push on the base when the brakes are engaged, \cref{fig:brakesON}, compared to when they are open, \cref{fig:brakesOFF}. Besides deformations of the base structure and tires at the contact point, the system with locked breaks does not move.

\begin{figure}[t]
   \centering
    \setlength\fwidth{0.8\linewidth}
    \setlength\fheight{0.4\linewidth}
    \input{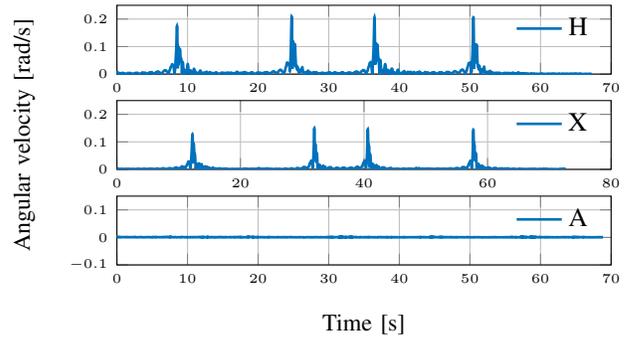}
\caption{Angular velocities of the base in different configurations, with one wheel raised \SI{15}{cm}, with the arm slowly spinning. Spikes can be observed for the \emph{H} and \emph{X-configuration} during the change of support polygon, while the \emph{A-configuration} is not perturbed by the motion of the ZMP}
\label{fig:angularVelocities}
\end{figure}

To validate the ST's contribution to stability, we raise one of the contact points of the robot, move its ZMP, and record the angular velocity of the base. We obtained a repeatable motion of the ZMP commanding the arm on top to rotate at low speed (making inertial load negligible) while horizontally stretched. The arm's mass caused the ZMP to move in a circle around the center of the base, as qualitatively shown in \cref{fig:overlayCOM}. We performed the tests raising only one wheel of \SI{15}{cm}, for all the robot configurations. \Cref{fig:angularVelocities} shows the time evolution of angular velocities, measured with an IMU connected to the base for the different configurations. It is possible to see how the ZMP motion affects the \emph{X-} and \emph{H-configuration}, causing changes of the STs and therefore spikes in the angular velocity. The \emph{A-configuration} is not perturbed by the motion of the ZMP.

\subsection{Driving experiments}
\label{ss:driving}
In a series of experiments, we command the robot to drive to different poses while transitioning between configurations.
The robot passes through a door and navigates on uneven ground.
The brakes are locked when the robot drives in \emph{H-configuration} and released for reconfiguration or driving in \emph{A-} or \emph{X-configuration}.
Linear interpolation between the current poses and the goal poses gives the reference base pose trajectories.
The integrated wheel odometry is passed to the MPC module as an observation for the base pose.
\subsubsection*{Result}
\begin{figure}
	\centering
	\includegraphics[width=1.0\linewidth]{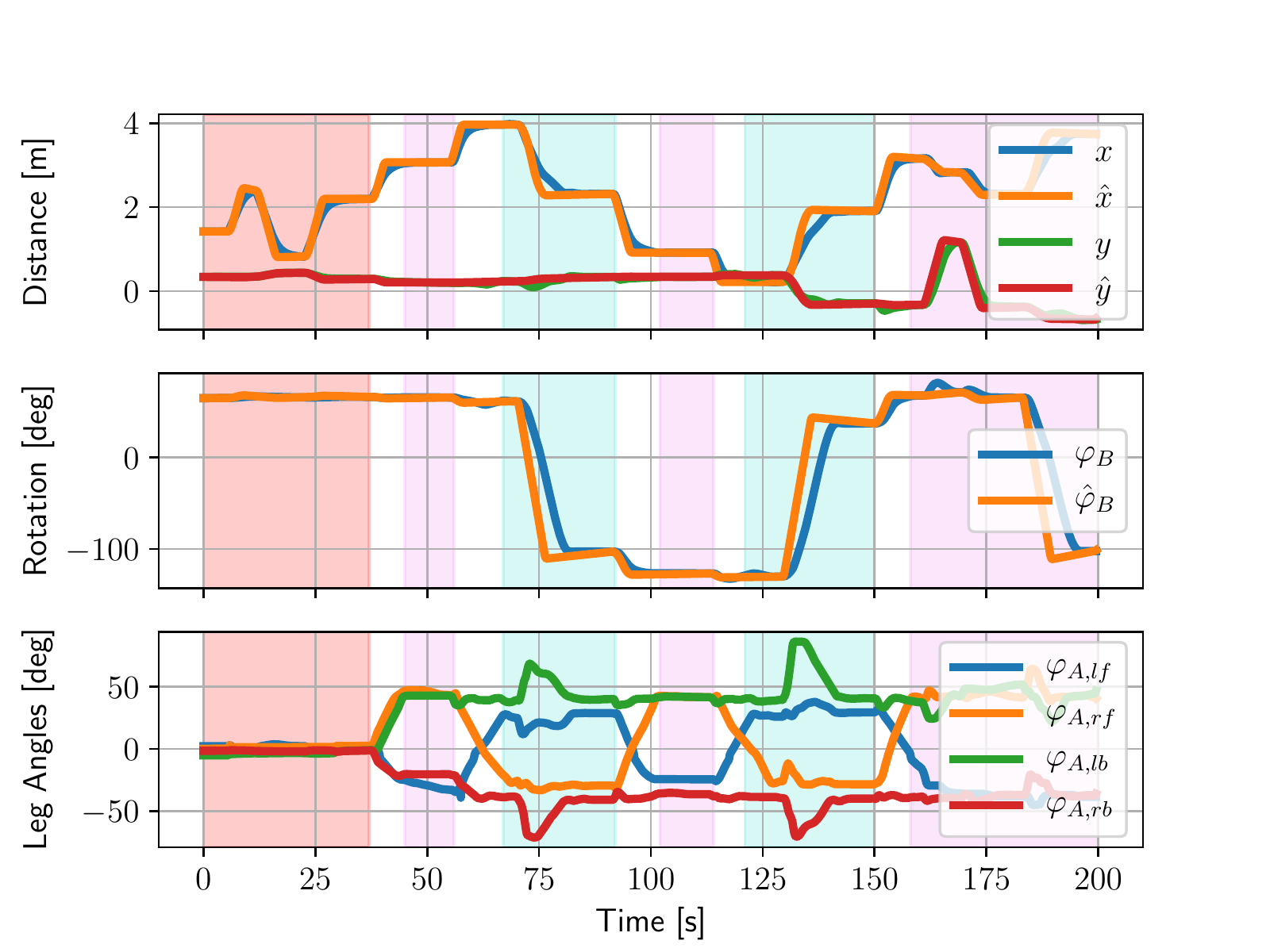}
	\caption{
	Driving with the reconfigurable base: The robot is commanded to follow a base trajectory during which the desired position, orientation, and base configuration changes. The controller can track the references. After the first section in H-configuration (red), the robot drives with released brakes and alternates between X-configuration (violet) and A-configuration (turquoise). Disturbances of the leg configurations can be observed, but the controller can always recover and steer back to the desired configuration.}
	\label{fig:driving_plot}
\end{figure}

\begin{figure*}[th]
	\centering
	\begin{subfigure}[t]{0.18\linewidth}
		\includegraphics[width=1.0\linewidth]{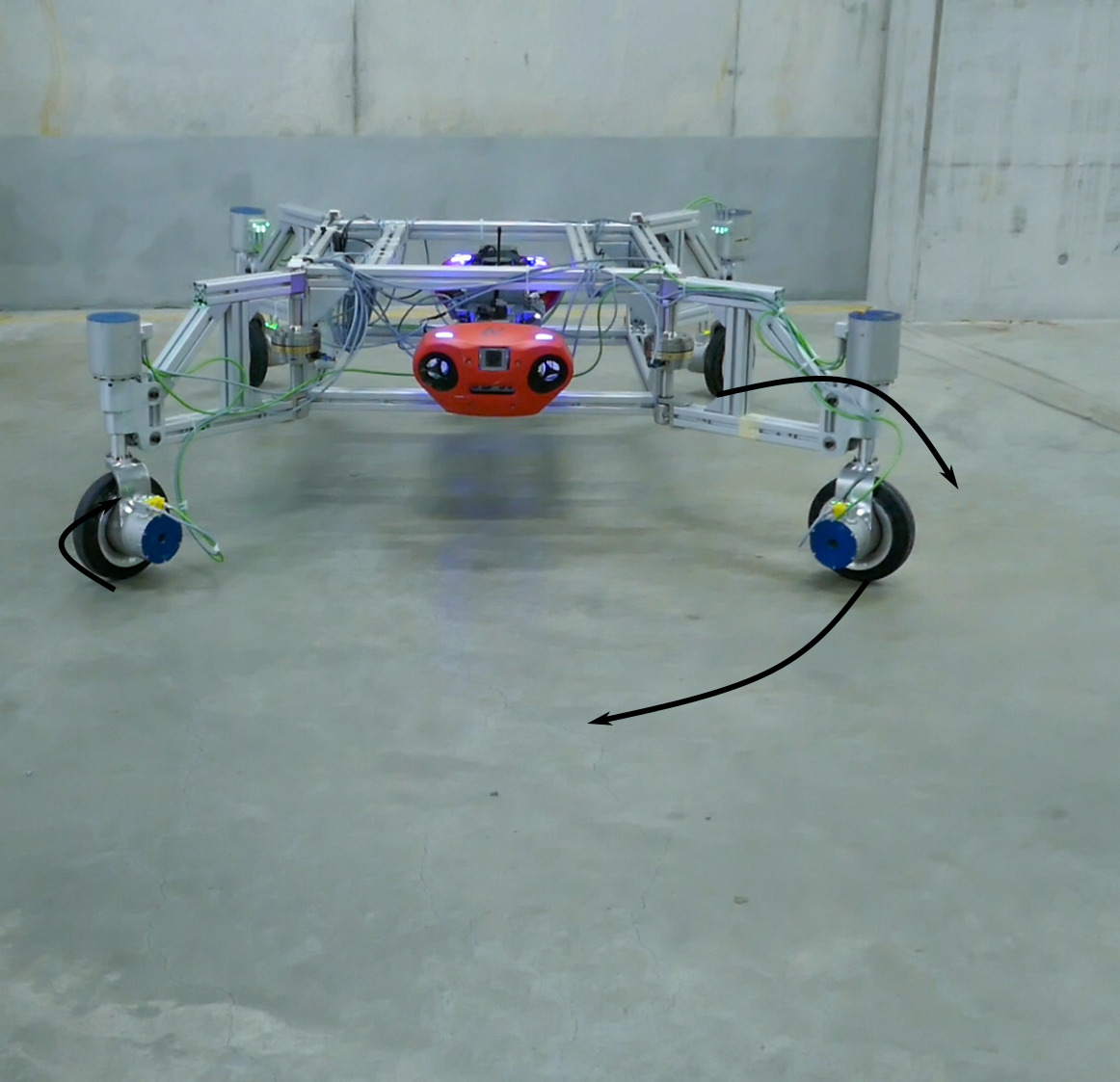}
		\caption{$t=\SI{0}{\second}$}
		\label{fig:driving_x}
	\end{subfigure}
	\hfill
	\begin{subfigure}[t]{0.18\linewidth}
		\includegraphics[width=1.0\linewidth]{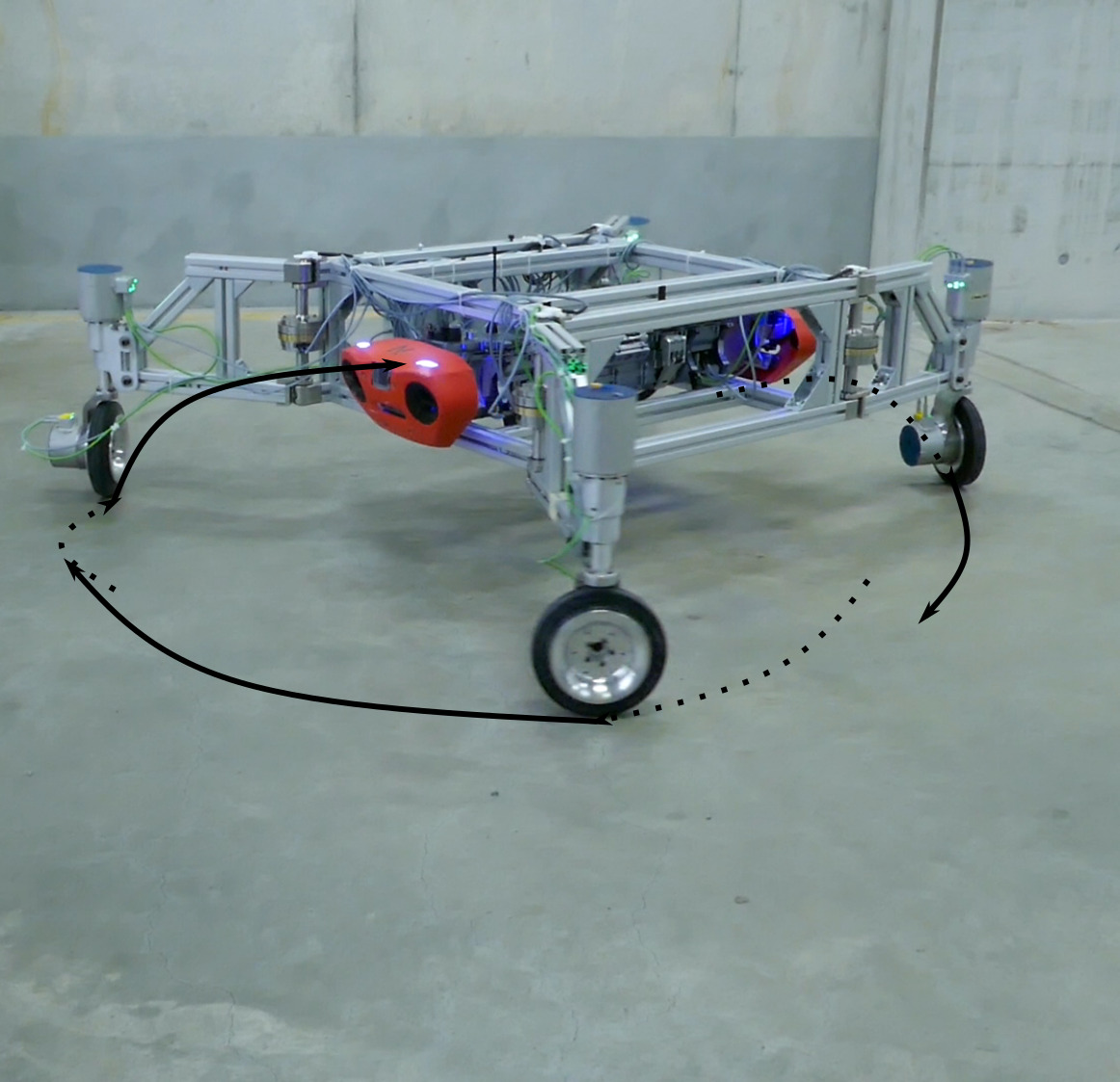}
		\caption{$t=\SI{5}{\second}$}
		\label{fig:driving_between_0}
	\end{subfigure}
	\hfill
	\begin{subfigure}[t]{0.18\linewidth}
		\includegraphics[width=1.0\linewidth]{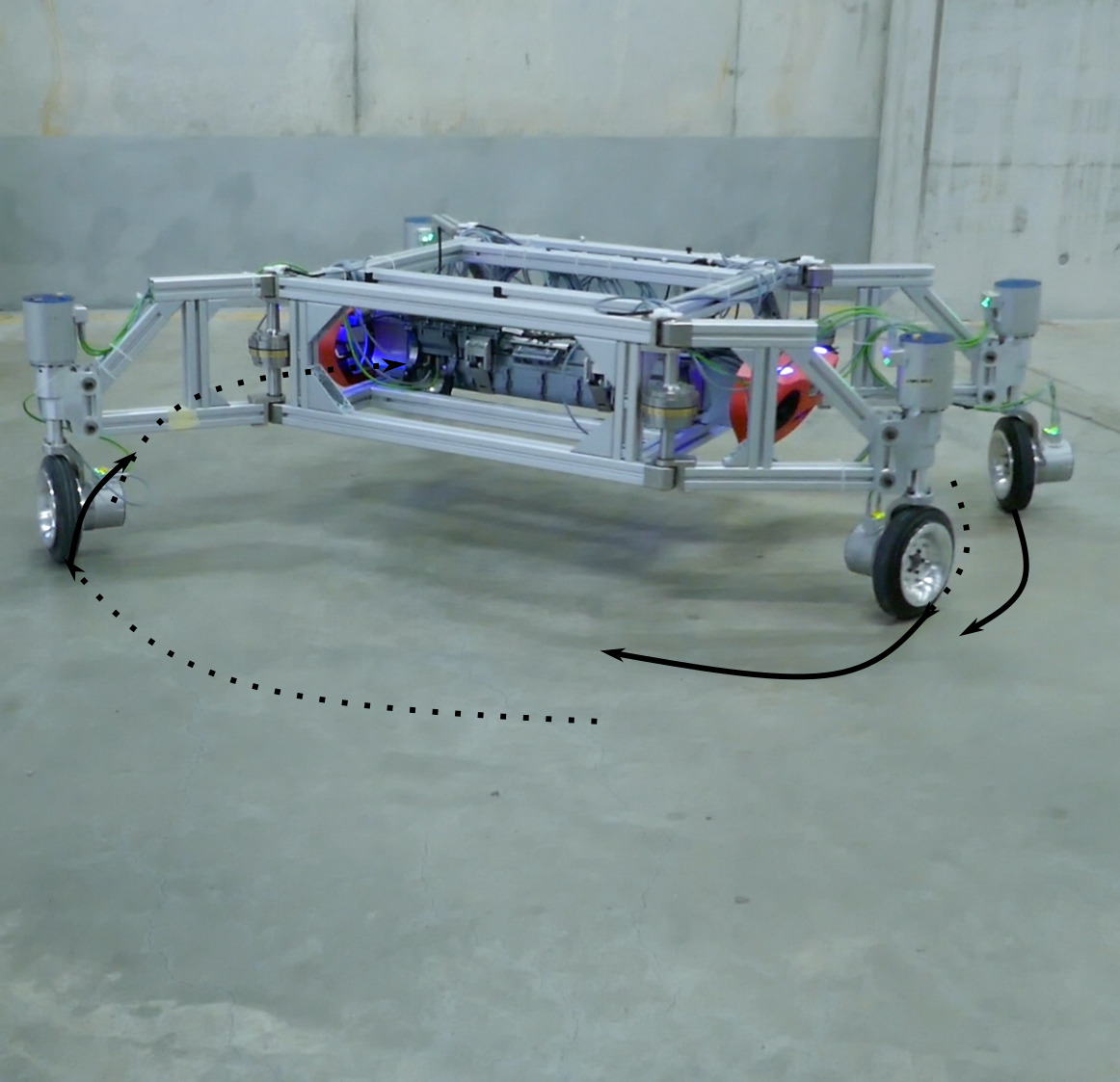}
		\caption{$t=\SI{8}{\second}$}
		\label{fig:driving_between_1}
	\end{subfigure}
	\hfill
	\begin{subfigure}[t]{0.18\linewidth}
		\includegraphics[width=1.0\linewidth]{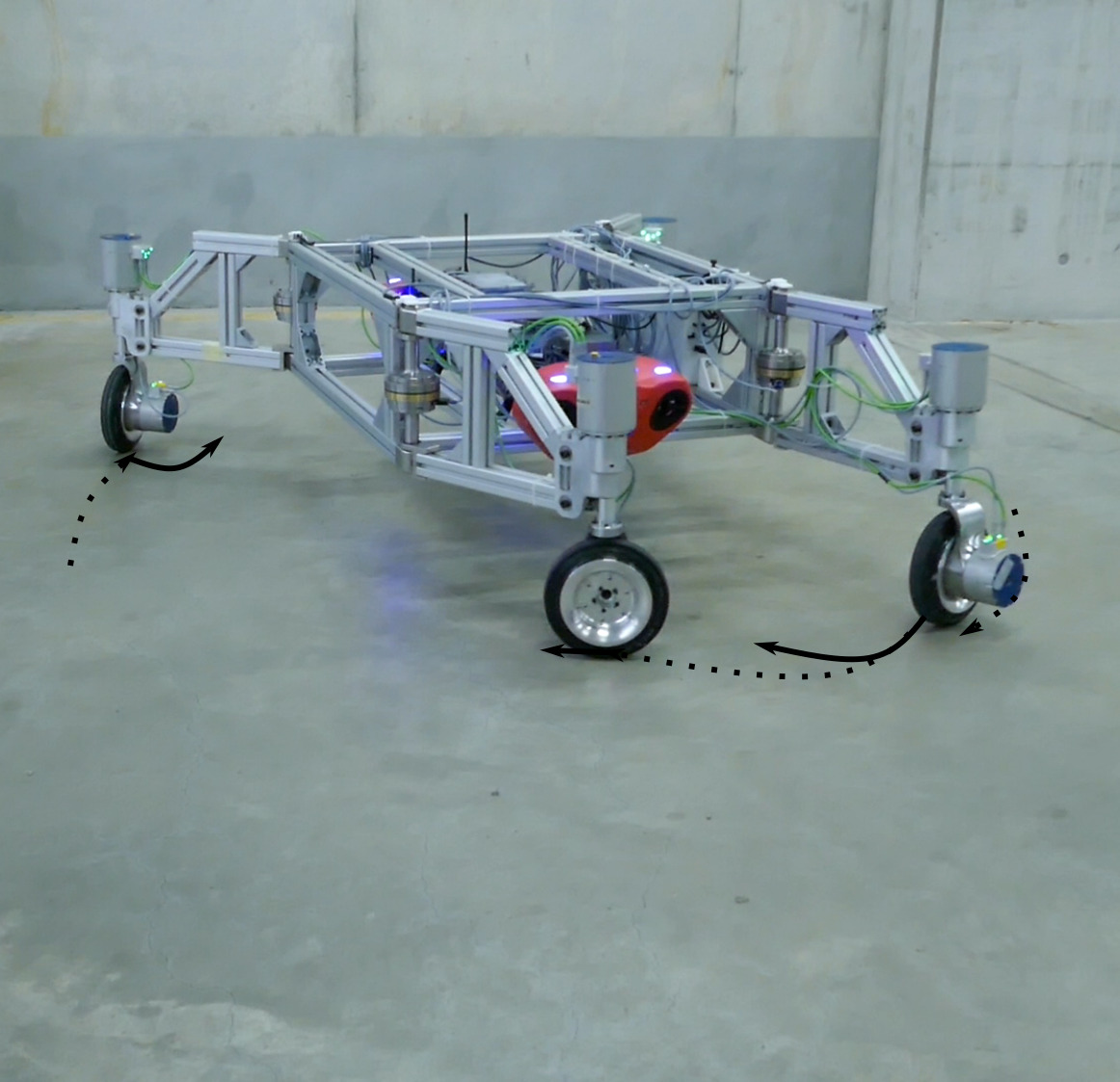}
		\caption{$t=\SI{10}{\second}$}
		\label{fig:driving_between_2}
	\end{subfigure}
		\hfill
	\begin{subfigure}[t]{0.18\linewidth}
		\includegraphics[width=1.0\linewidth]{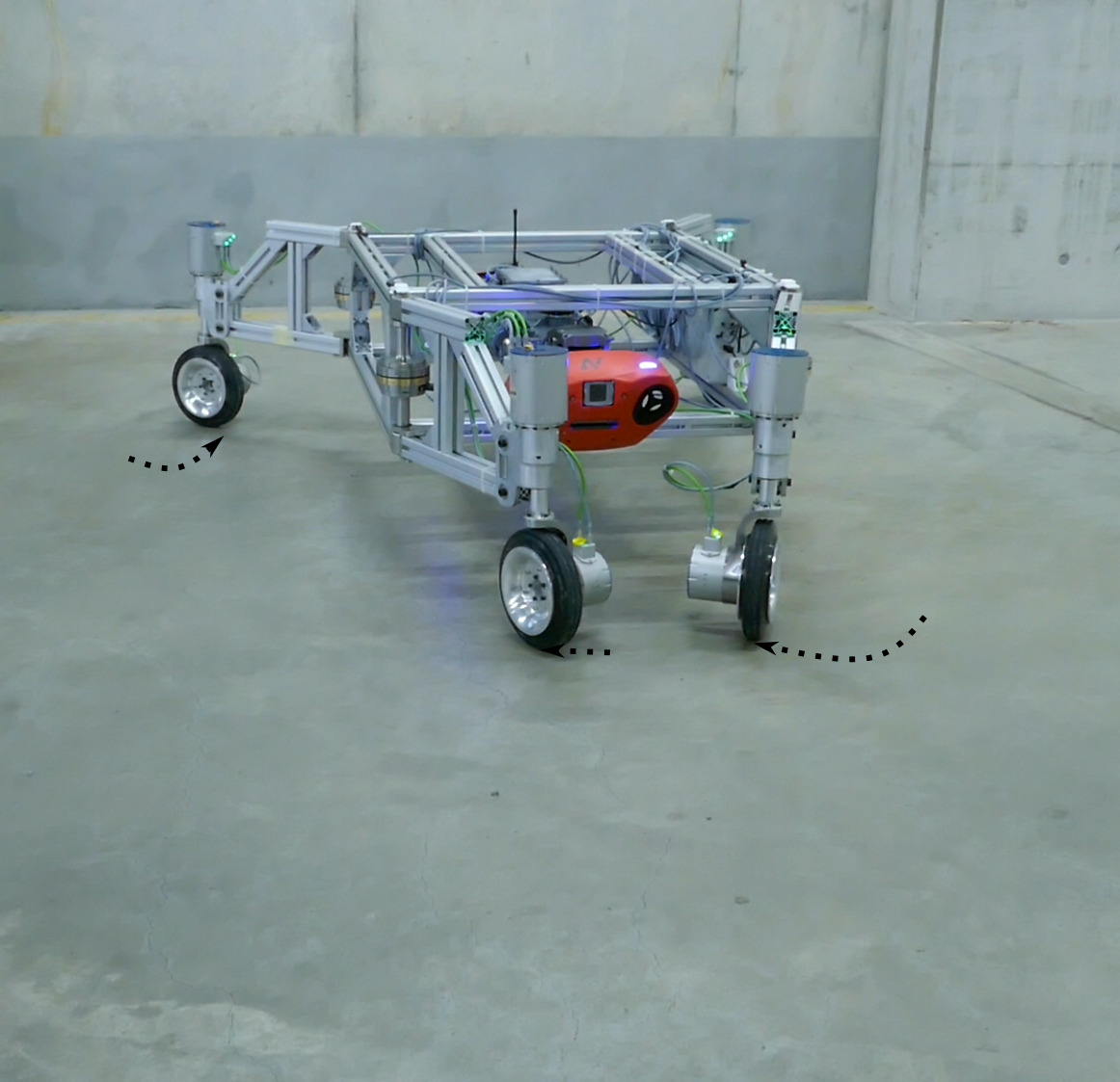}
		\caption{$t=\SI{18}{\second}$}
		\label{fig:driving_a}
	\end{subfigure}
	\begin{subfigure}[t]{0.18\linewidth}
		\includegraphics[width=1.0\linewidth]{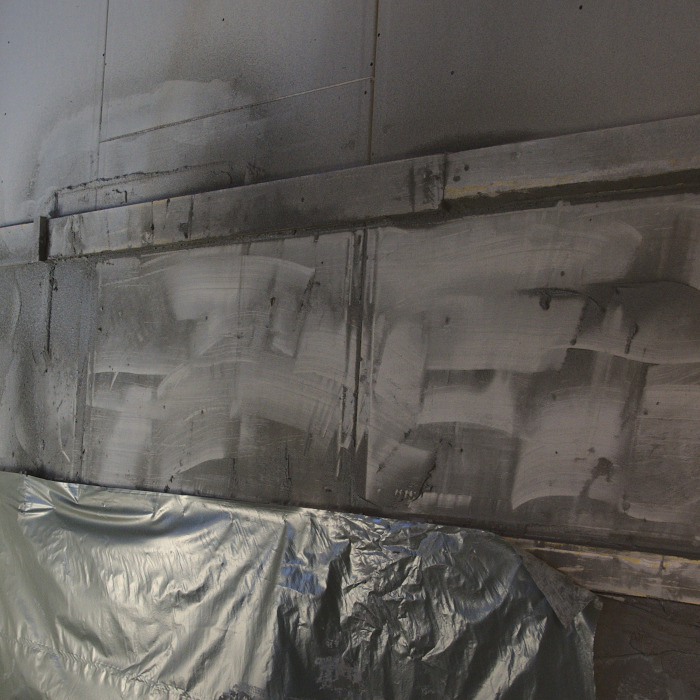}
		\caption{Initial surface}
		\label{fig:initial_wall}
	\end{subfigure}
	\hfill
	\begin{subfigure}[t]{0.18\linewidth}
		\includegraphics[width=1.0\linewidth]{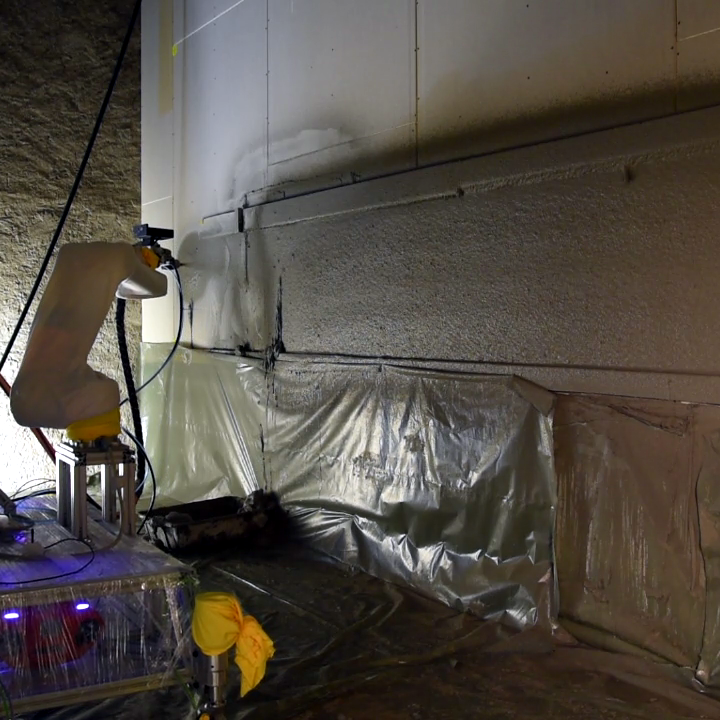}
		\caption{$t=t_0$}
		\label{fig:plastering_1}
	\end{subfigure}
	\hfill
	\begin{subfigure}[t]{0.18\linewidth}
		\includegraphics[width=1.0\linewidth]{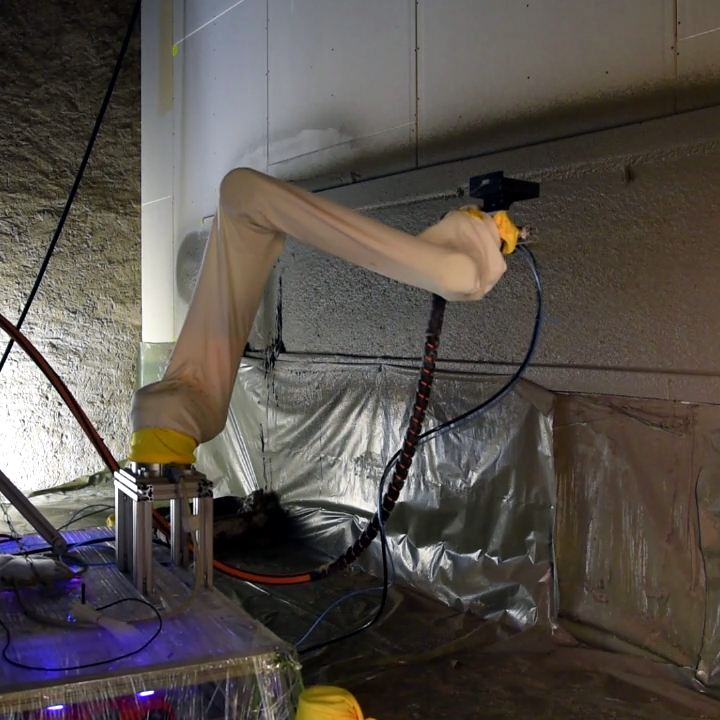}
		\caption{$t=t_0+\SI{5}{\second}$}
		\label{fig:plastering_2}
	\end{subfigure}
	\hfill
 	\begin{subfigure}[t]{0.18\linewidth}
 		\includegraphics[width=1.0\linewidth]{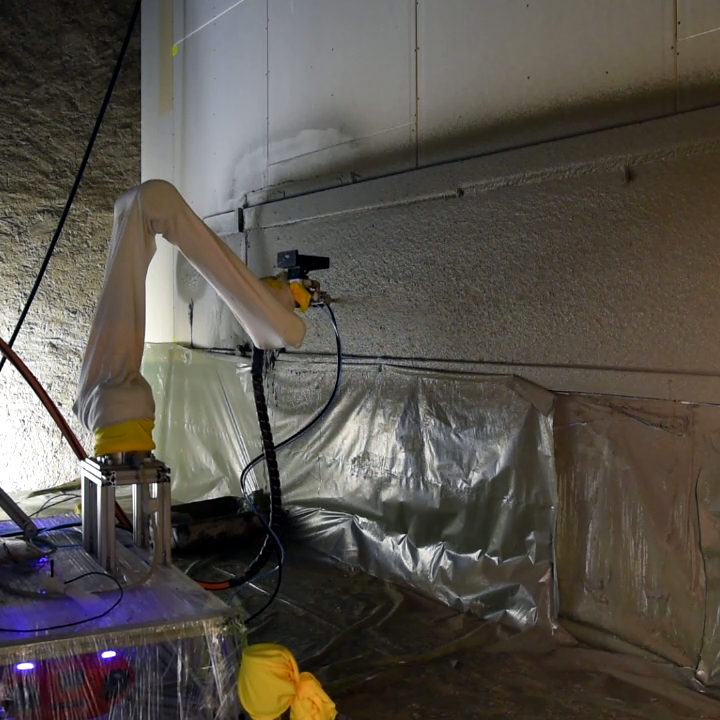}
 		\caption{$t=t_0+\SI{10}{\second}$}
 		\label{fig:plastering_3}
 	\end{subfigure}
 	\hfill
 	\begin{subfigure}[t]{0.18\linewidth}
		\includegraphics[width=1.0\linewidth]{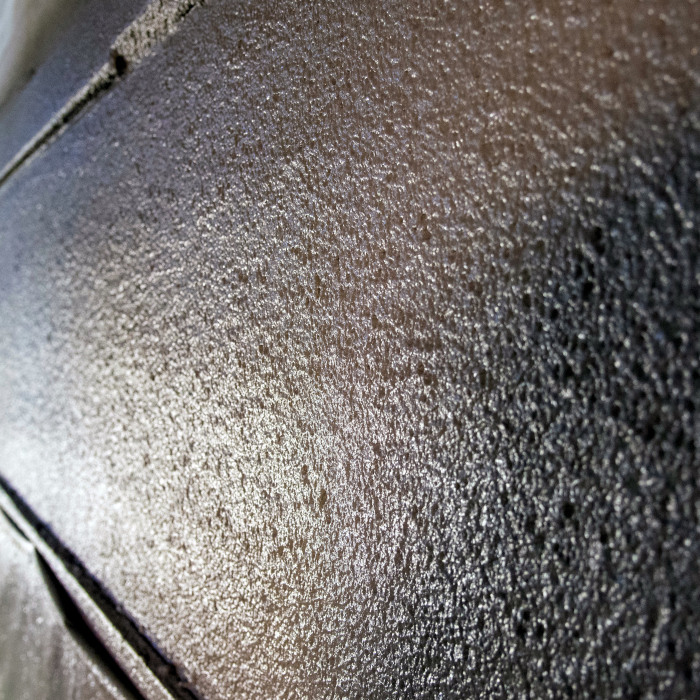}
		\caption{Finished surface}
		\label{fig:final_plaster}
	\end{subfigure}
	\caption{Top row: Stills of the reconfiguration process. The robot transitions between X and A configuration, while turning by $\SI{180}{\degree}$.
	Solid arrows indicate the motion of the wheels from the current frame to the next frame.
	Dotted arrows show the movement of the wheels from the previous frame to the current frame.\\
	Bottom row: Stills of automated plastering process in \crefrange{fig:plastering_1}{fig:plastering_3}.
	\Cref{fig:initial_wall} shows the wall before the plastering process. \Cref{fig:final_plaster} shows the final plaster coating. Photo credits: Selen Ercan Jenny
	}
	\label{fig:driving_reconfigure}
\end{figure*}

In the accompanying video \footnote{\url{https://youtu.be/qBY4zovf2vo}}, the results of the driving tests are presented.
\Cref{fig:driving_reconfigure} shows how the robot successfully transitions between \emph{X-} and \emph{A-configuration} while rotating by \SI{180}{\degree}.
We display the driving trajectory of one experiment in \cref{fig:driving_plot}.
The position and orientation reference, the recorded robot pose, and the leg angles are plotted over time.
We can see that the base poses can be tracked and that the robot can simultaneously move and change its configuration.
When the robot drives with its brakes released, leg angles sometimes deviate from the desired value. This can be caused by the solver prioritizing base tracking or actuation costs over leg configuration tracking. In all tested cases, the robot is able to quickly recover from the deviation and return to the desired configuration. The brakes can stay closed in application cases where the base orientation has to be fixed.
\subsection{Whole-Body End-Effector Tracking}
\label{ss:ee_tracking}
Whole-body end-effector tracking with a mobile robot aims to perform continuous manipulation actions in a workspace that is larger than the reach of a fixed base manipulator.
The MPC should compute and track motion plans that control all degrees of freedom to follow the desired end-effector trajectory optimally.
We mount a UR10 manipulator on the mobile base and attach a reflective prism to its end-effector to evaluate the tracking performance.
The prism is tracked with a total station that provides absolute ground truth position information.
The robot is commanded to track a sinusoidal path in the air. The reference path is $\SI{20.53}{\meter}$ long. We use integrated wheel odometry for the base state observation.
\subsubsection*{Results}
\begin{figure}
	\centering
	\includegraphics[width=1.0\linewidth]{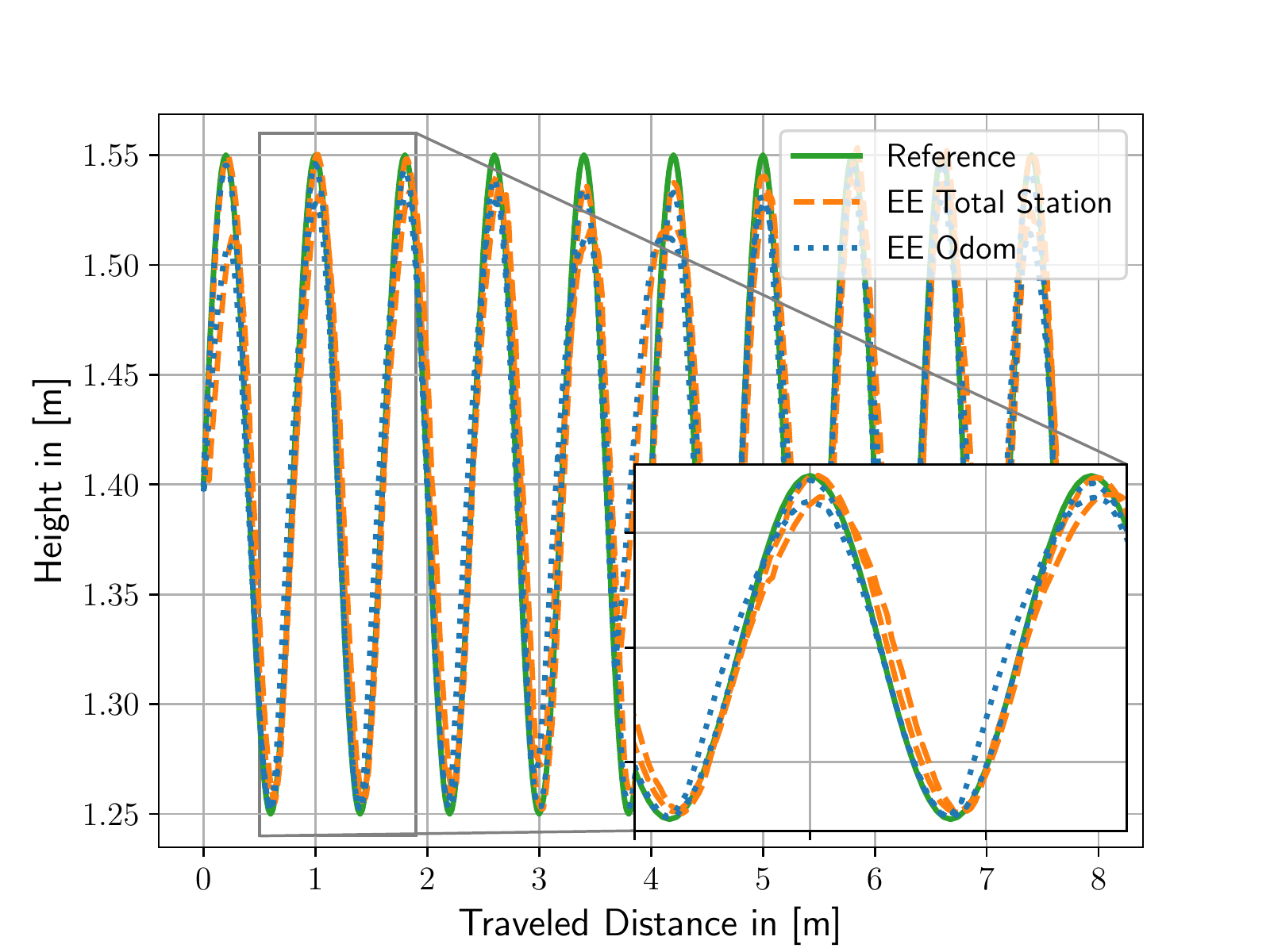}
	\caption{End-effector tracking experiments: The robot tracks a sinusoidal reference path (green, solid). The end-effector position in odometry frame is shown in dotted blue. Ground truth position information was recorded with a total station (orange, dashed).}
	\label{fig:end_effector_tracking}
\end{figure}

\Cref{fig:end_effector_tracking} shows the tracking accuracy of the end-effector in the experiment.
The mean average error (MAE) between the reference path and the end-effector position in the odometry frame is $\SI{1.67}{\centi\meter}$.
This tracking performance is sufficient for the application use case presented in \cref{ss:plastering}.
The peaks of the reference path can not always be reached, since the arm operates close to its maximum reach.
The base odometry drifts $\SI{16.1}{\centi\meter}$ away from the total station ground truth throughout $\SI{15.96}{\meter}$ of base traveling.\\
For higher absolute accuracy, a LIDAR or camera-based localization system should be integrated. The global base pose estimate could be passed to the MPC as a drop-in replacement for the integrated wheel odometry.
\subsection{Construction Demo}
\label{ss:plastering}
The mobile manipulator is used for automated plaster deposition to achieve a flat wall finish. For that application, the mobile platform is equipped with a custom plaster spray gun attached to a UR10 robotic arm. The construction site poses an unstructured environment with uneven floors. The small obstacles on the ground, such as cables, need to be driven over and can stop wheels at low speeds creating disruptions to a planned motion. The MPC should overcome such deviations to the initial trajectory by replanning a motion sequence accounting for a present disturbance. Using the whole-body end-effector tracking, we command a spray gun to follow a trajectory consisting of 5 connected horizontal lines. During the experiment, we aim to cover in plaster a part of the wall with the following dimensions: $\SI{3.50}{\meter}$ length and $\SI{0.75}{\meter}$ height. To achieve even coating, the spacing between horizontal lines is equal to $\SI{0.15}{\meter}$ with commanded speeds of $\SI{0.7}{\meter/\second}$. The trajectory is repeated ten times resulting in a total end-effector path length of around $\SI{180}{\meter}$. Wheel odometry is employed for base state observation.\\
For more details on the robotic spraying process, please refer to prior work of Ercan Jenny et al. \cite{ercanjennyContinuousMobileThinLayer2023}.
\subsubsection*{Result}
\Crefrange{fig:initial_wall}{fig:final_plaster} shows the wall before plastering, stills from the plastering process, and the surface of the finished wall.
The proposed method copes well with disturbances resulting from an unstructured environment. The platform can repeat the path while overcoming the aforementioned small obstacles. Benefiting from the controller's high repeatability, the system successfully fabricates a flat plaster surface on the wall.
\section{CONCLUSIONS \& FUTURE WORK}
\label{sec:conclusions}
We have shown how a reconfigurable mobile base can satisfy the diverse requirements to be deployed on a construction site. The robot successfully passes through doors in the narrow configuration. By assuming a specific configuration for manipulation, the roll and pitch velocity of the base can be reduced by order of magnitude compared to the default configuration.
The proposed MPC enables the robot to switch between configurations autonomously while tracking base navigation references.
It also allows for precise end-effector tracking with whole-body motion, which has been shown in lab experiments and a construction demonstration.\\
For future work, perception should be integrated into the motion planning pipeline. Autonomous reconfiguration could be used for collision avoidance during navigation planning. Experiments need to show whether it is sufficient to integrate collision avoidance on the MPC level or if a hierarchical structure with a sampling-based planner combined with an MPC tracking module is necessary.\\
The MPC currently uses a kinematic model and neglects the dynamic effects of base and arm motion. Kinematic planning has been sufficient for our application, but a dynamic system model should be considered for scenarios that require faster movement.
Another interesting direction is to explore different morphologies for the base by changing the number of reconfigurable legs or replacing the passive hinge joints with prismatic joints.
\bibliographystyle{IEEEtran}
\bibliography{references}

\begin{thebibliography}{10}
\providecommand{\url}[1]{#1}
\csname url@samestyle\endcsname
\providecommand{\newblock}{\relax}
\providecommand{\bibinfo}[2]{#2}
\providecommand{\BIBentrySTDinterwordspacing}{\spaceskip=0pt\relax}
\providecommand{\BIBentryALTinterwordstretchfactor}{4}
\providecommand{\BIBentryALTinterwordspacing}{\spaceskip=\fontdimen2\font plus
\BIBentryALTinterwordstretchfactor\fontdimen3\font minus
  \fontdimen4\font\relax}
\providecommand{\BIBforeignlanguage}[2]{{%
\expandafter\ifx\csname l@#1\endcsname\relax
\typeout{** WARNING: IEEEtran.bst: No hyphenation pattern has been}%
\typeout{** loaded for the language `#1'. Using the pattern for}%
\typeout{** the default language instead.}%
\else
\language=\csname l@#1\endcsname
\fi
#2}}
\providecommand{\BIBdecl}{\relax}
\BIBdecl

\bibitem{TheMcKinsey}
\BIBentryALTinterwordspacing
``{The impact and opportunities of automation in construction | McKinsey}.''
  [Online]. Available:
  \url{https://www.mckinsey.com/business-functions/operations/our-insights/the-impact-and-opportunities-of-automation-in-construction}
\BIBentrySTDinterwordspacing

\bibitem{suva2020}
\BIBentryALTinterwordspacing
{Koordinationsgruppe f{\"{u}}r die Statistik der Unfallversicherung UVG
  (KSUV)}, ``{Unfallstatistik UVG 2020},'' KSUV, Tech. Rep., 2020. [Online].
  Available: \url{www.unfallstatistik.ch/UnfallstatistikUVG2020}
\BIBentrySTDinterwordspacing

\bibitem{hack2017mesh}
N.~Hack, T.~Wangler, J.~Mata-Falc{\'o}n, K.~D{\"o}rfler, N.~Kumar, A.~N.
  Walzer, K.~Graser, L.~Reiter, H.~Richner, J.~Buchli \emph{et~al.}, ``Mesh
  mould: an on site, robotically fabricated, functional formwork,'' in
  \emph{Second Concrete Innovation Conference (2nd CIC), Paper}, no.~19, 2017.

\bibitem{helm2012mobile}
V.~Helm, S.~Ercan, F.~Gramazio, and M.~Kohler, ``Mobile robotic fabrication on
  construction sites: Dimrob,'' in \emph{2012 IEEE/RSJ International Conference
  on Intelligent Robots and Systems}.\hskip 1em plus 0.5em minus 0.4em\relax
  IEEE, 2012, pp. 4335--4341.

\bibitem{WheeledCompany}
\BIBentryALTinterwordspacing
``{Wheeled Platform | AMBOT | American Robot Company}.'' [Online]. Available:
  \url{http://www.ambot.com/ip-wheel.shtml#scroll-grp4400}
\BIBentrySTDinterwordspacing

\bibitem{Halme2000HybridMachine}
A.~Halme, I.~Lepp{\"{a}}nen, S.~Salmi, and S.~Yl{\"{o}}nen, ``{Hybrid
  locomotion of a wheel-legged machine},'' in \emph{3rd Int. Conference on
  Climbing and Walking Robots}, 2000.

\bibitem{Milliken1988ActiveSuspension}
W.~F. Milliken, ``{Active Suspension},'' in \emph{SAE Technical Paper 880799},
  4 1988.

\bibitem{2017PatrickGreat}
\BIBentryALTinterwordspacing
``{Patrick Head explains why the 1992 Williams FW14B was great},'' 12 2017.
  [Online]. Available:
  \url{https://www.autosport.com/f1/news/patrick-head-explains-why-the-1992-williams-fw14b-was-great-4989593/4989593/}
\BIBentrySTDinterwordspacing

\bibitem{Pioneer3-DX}
\BIBentryALTinterwordspacing
``{Pioneer 3-DX}.'' [Online]. Available: \url{www.mobilerobots.com}
\BIBentrySTDinterwordspacing

\bibitem{RB-1Robotnik}
\BIBentryALTinterwordspacing
``{RB-1 BASE - Warehouse Logistics Robot | Robotnik{\textregistered}}.''
  [Online]. Available:
  \url{https://robotnik.eu/products/mobile-robots/rb-1-base-en/}
\BIBentrySTDinterwordspacing

\bibitem{OutriggersCrane}
\BIBentryALTinterwordspacing
``{Outriggers and Stabilizers – Auto Crane}.'' [Online]. Available:
  \url{https://autocrane.us/product/outriggers-stabilizers/}
\BIBentrySTDinterwordspacing

\bibitem{JanBirgerPalmcrantz1973SupportCranes}
{Jan Birger Palmcrantz}, ``{Support Legs of Mobile Cranes},'' 1973.

\bibitem{RemoteProducts}
\BIBentryALTinterwordspacing
``{Remote controlled demolition robots | Husqvarna Construction Products}.''
  [Online]. Available:
  \url{https://www.husqvarnacp.com/int/machines/demolition-robots/}
\BIBentrySTDinterwordspacing

\bibitem{Mandrovskiy2018OptimizingStability}
K.~P. Mandrovskiy and Y.~I. Tyurin, ``{Optimizing the Support Polygon of a
  Wheeled Excavator in Terms of Stability},'' \emph{Russian Engineering
  Research}, vol.~38, no.~1, 1 2018.

\bibitem{Kamedula2020ReactiveRobot}
M.~Kamedula and N.~G. Tsagarakis, ``{Reactive Support Polygon Adaptation for
  the Hybrid Legged-Wheeled CENTAURO Robot},'' \emph{IEEE Robotics and
  Automation Letters}, vol.~5, no.~2, 4 2020.

\bibitem{Hutter2016ANYmalRobot}
M.~Hutter, C.~Gehring, D.~Jud, A.~Lauber, C.~D. Bellicoso, V.~Tsounis,
  J.~Hwangbo, K.~Bodie, P.~Fankhauser, M.~Bloesch, R.~Diethelm, S.~Bachmann,
  A.~Melzer, and M.~Hoepflinger, ``{ANYmal - a highly mobile and dynamic
  quadrupedal robot},'' in \emph{2016 IEEE/RSJ International Conference on
  Intelligent Robots and Systems (IROS)}.\hskip 1em plus 0.5em minus
  0.4em\relax IEEE, 10 2016.

\bibitem{StretchDynamics}
\BIBentryALTinterwordspacing
``{Stretch | Boston Dynamics}.'' [Online]. Available:
  \url{https://www.bostondynamics.com/stretch}
\BIBentrySTDinterwordspacing

\bibitem{BostonPalletizing}
\BIBentryALTinterwordspacing
``{Boston Dynamics' Stretch robot handles truck unloading {\&} palletizing}.''
  [Online]. Available:
  \url{https://www.therobotreport.com/boston-dynamics-stretch-robot-truck-unloading-palletizing/}
\BIBentrySTDinterwordspacing

\bibitem{Fu2014TheROAMeR}
Q.~Fu, X.~Zhou, and V.~Krovi, ``{The Reconfigurable Omnidirectional Articulated
  Mobile Robot (ROAMeR)},'' \emph{Springer Tracts in Advanced Robotics 79},
  2014.

\bibitem{Yun2021DevelopmentEnvironments}
S.-H. Yun, J.~Park, J.~Seo, and Y.-J. Kim, ``{Development of an Agile
  Omnidirectional Mobile Robot With GRF Compensated Wheel-leg Mechanisms for
  Human Environments},'' \emph{IEEE Robotics and Automation Letters}, vol.~6,
  no.~4, 10 2021.

\bibitem{anymal_on_wheels}
M.~Bjelonic, C.~D. Bellicoso, Y.~de~Viragh, D.~Sako, F.~D. Tresoldi,
  F.~Jenelten, and M.~Hutter, ``Keep rollin’—whole-body motion control and
  planning for wheeled quadrupedal robots,'' \emph{IEEE Robotics and Automation
  Letters}, vol.~4, no.~2, pp. 2116--2123, 2019.

\bibitem{burget_whole-body_2013}
F.~Burget, A.~Hornung, and M.~Bennewitz, ``Whole-body motion planning for
  manipulation of articulated objects,'' in \emph{2013 {IEEE} {International}
  {Conference} on {Robotics} and {Automation}}, May 2013, pp. 1656--1662, iSSN:
  1050-4729.

\bibitem{raghavan_variable_2019}
V.~S. Raghavan, D.~Kanoulas, A.~Laurenzi, D.~G. Caldwell, and N.~G. Tsagarakis,
  ``Variable {Configuration} {Planner} for {Legged}-{Rolling} {Obstacle}
  {Negotiation} {Locomotion}: {Application} on the {Centauro} {Robot},'' in
  \emph{2019 {IEEE}/{RSJ} {International} {Conference} on {Intelligent}
  {Robots} and {Systems} ({IROS})}, Nov. 2019, pp. 4738--4745, iSSN: 2153-0866.

\bibitem{honerkamp2021learning}
D.~Honerkamp, T.~Welschehold, and A.~Valada, ``Learning kinematic feasibility
  for mobile manipulation through deep reinforcement learning,'' \emph{IEEE
  Robotics and Automation Letters (RA-L)}, 2021.

\bibitem{alma_learning}
Y.~Ma, F.~Farshidian, T.~Miki, J.~Lee, and M.~Hutter, ``Combining
  learning-based locomotion policy with model-based manipulation for legged
  mobile manipulators,'' \emph{IEEE Robotics and Automation Letters}, vol.~7,
  no.~2, pp. 2377--2384, 2022.

\bibitem{sun2022fully}
C.~Sun, J.~Orbik, C.~M. Devin, B.~H. Yang, A.~Gupta, G.~Berseth, and S.~Levine,
  ``Fully autonomous real-world reinforcement learning with applications to
  mobile manipulation,'' in \emph{Conference on Robot Learning}.\hskip 1em plus
  0.5em minus 0.4em\relax PMLR, 2022, pp. 308--319.

\bibitem{alma_mpc}
J.-P. Sleiman, F.~Farshidian, M.~V. Minniti, and M.~Hutter, ``A unified mpc
  framework for whole-body dynamic locomotion and manipulation,'' \emph{IEEE
  Robotics and Automation Letters}, vol.~6, no.~3, pp. 4688--4695, 2021.

\bibitem{giftthaler_efficient_2017}
M.~Giftthaler, F.~Farshidian, T.~Sandy, L.~Stadelmann, and J.~Buchli,
  ``Efficient kinematic planning for mobile manipulators with non-holonomic
  constraints using optimal control,'' in \emph{2017 {IEEE} {International}
  {Conference} on {Robotics} and {Automation} ({ICRA})}, May 2017, pp.
  3411--3417.

\bibitem{pankert2020perceptive}
J.~Pankert and M.~Hutter, ``Perceptive model predictive control for continuous
  mobile manipulation,'' \emph{IEEE Robotics and Automation Letters}, vol.~5,
  no.~4, pp. 6177--6184, 2020.

\bibitem{zmp}
M.~Vukobratovic and B.~Borovac, ``Zero-moment point - thirty five years of its
  life.'' \emph{I. J. Humanoid Robotics}, vol.~1, pp. 157--173, 03 2004.

\bibitem{shigemi2018asimo}
S.~Shigemi, A.~Goswami, and P.~Vadakkepat, ``Asimo and humanoid robot research
  at honda,'' \emph{Humanoid robotics: A reference}, pp. 55--90, 2018.

\bibitem{kupplungshersteller}
\BIBentryALTinterwordspacing
``Mönninghoff brakes.'' [Online]. Available:
  \url{https://moenninghoff.de/en/products/brakes/electromagnetic-tooth-brakes/typ-560-elektromagnet-zahnhaltebremse}
\BIBentrySTDinterwordspacing

\bibitem{seeedstudio}
\BIBentryALTinterwordspacing
``Grove - 12 bit magnetic rotary position sensor(as5600).'' [Online].
  Available:
  \url{https://wiki.seeedstudio.com/Grove-12-bit-Magnetic-Rotary-Position-Sensor-AS5600/}
\BIBentrySTDinterwordspacing

\bibitem{farshidian2017efficient}
F.~Farshidian, M.~Neunert, A.~W. Winkler, G.~Rey, and J.~Buchli, ``An efficient
  optimal planning and control framework for quadrupedal locomotion,'' in
  \emph{2017 IEEE International Conference on Robotics and Automation
  (ICRA)}.\hskip 1em plus 0.5em minus 0.4em\relax IEEE, 2017, pp. 93--100.

\bibitem{ros_control}
\BIBentryALTinterwordspacing
S.~Chitta, E.~Marder-Eppstein, W.~Meeussen, V.~Pradeep,
  A.~Rodr{\'i}guez~Tsouroukdissian, J.~Bohren, D.~Coleman, B.~Magyar,
  G.~Raiola, M.~L{\"u}dtke, and E.~Fern{\'a}ndez~Perdomo, ``ros\_control: A
  generic and simple control framework for ros,'' \emph{The Journal of Open
  Source Software}, 2017. [Online]. Available:
  \url{http://www.theoj.org/joss-papers/joss.00456/10.21105.joss.00456.pdf}
\BIBentrySTDinterwordspacing

\bibitem{FourWheelSteering}
\BIBentryALTinterwordspacing
``four\_wheel\_steering\_controller - {ROS} {Wiki} - last accessed:
  2022-02-19.'' [Online]. Available:
  \url{http://wiki.ros.org/four_wheel_steering_controller}
\BIBentrySTDinterwordspacing

\bibitem{ercanjennyContinuousMobileThinLayer2023}
\BIBentryALTinterwordspacing
S.~Ercan~Jenny, L.~L. Pietrasik, E.~Sounigo, P.-H. Tsai, F.~Gramazio,
  M.~Kohler, E.~Lloret-Fritschi, and M.~Hutter,
  ``\BIBforeignlanguage{en}{Continuous {Mobile} {Thin}-{Layer} {On}-{Site}
  {Printing}},'' \emph{\BIBforeignlanguage{en}{Automation in Construction}},
  vol. 146, p. 104634, Feb. 2023. [Online]. Available:
  \url{https://www.sciencedirect.com/science/article/pii/S0926580522005040}
\BIBentrySTDinterwordspacing

\end{thebibliography}
\end{document}